\documentclass[10pt,twocolumn,letterpaper]{article}

\usepackage[pagenumbers]{cvpr} 

\definecolor{cvprblue}{rgb}{0.21,0.49,0.74}
\usepackage[pagebackref,breaklinks,colorlinks,allcolors=cvprblue]{hyperref}

\usepackage{comment}

\usepackage[ruled,vlined]{algorithm2e}
\usepackage{amsmath}
\usepackage{amssymb}

\usepackage{multirow} 
\usepackage{multicol}
\usepackage{booktabs}

\usepackage{paralist}
\usepackage{subcaption}

\definecolor{lightblue}{HTML}{C7DEE9}

\def\paperID{} 
\def\confName{CVPR}
\def\confYear{2026}

\title{Dynamic Dual Buffer with Divide-and-Conquer Strategy for\\ Online Continual Learning}

\author{Congren Dai\textsuperscript{1} \quad Huichi Zhou\textsuperscript{2} \quad Jiahao Huang\textsuperscript{2} \quad Zhenxuan Zhang\textsuperscript{2} \\ 
Fanwen Wang\textsuperscript{2} \quad Yijian Gao\textsuperscript{1} \quad Guang Yang\textsuperscript{2} \quad Fei Ye\textsuperscript{3,\textdagger} 
\vspace{5pt}
\\
\textsuperscript{1}Department of Computing, Imperial College London, United Kingdom \\
\textsuperscript{2}Department of Bioengineering, Imperial College London, United Kingdom \\
\textsuperscript{3}School of Information and Software Engineering, \\
University of Electronic Science and Technology of China, China \\
\vspace{5pt}
\small{
\texttt{congren.dai@imperial.ac.uk, fy689@uestc.edu.cn}
}
}

\begin{document}
\maketitle

\begingroup
\renewcommand\thefootnote{\textdagger}
\footnotetext{Corresponding author.}
\endgroup

\newcommand{\model}{ODEDM}

\begin{abstract}
Online Continual Learning (OCL) involves sequentially arriving data and is particularly challenged by catastrophic forgetting, which significantly impairs model performance. To address this issue, we introduce a novel framework, Online Dynamic Expandable Dual Memory (ODEDM), that integrates a short-term memory for fast memory and a long-term memory structured into sub-buffers anchored by cluster prototypes, enabling the storage of diverse and category-specific samples to mitigate forgetting. We propose a novel $K$-means-based strategy for prototype identification and an optimal transport-based mechanism to retain critical samples, prioritising those exhibiting high similarity to their corresponding prototypes. This design preserves semantically rich information. Additionally, we propose a Divide-and-Conquer (DAC) optimisation strategy that decomposes memory updates into subproblems, thereby reducing computational overhead. ODEDM functions as a plug-and-play module that can be seamlessly integrated with existing rehearsal-based approaches. Experimental results under both standard and imbalanced OCL settings show that ODEDM consistently achieves state-of-the-art performance across multiple datasets, delivering substantial improvements over the DER family as well as recent methods such as VR-MCL and POCL.
\end{abstract}

\vspace{-10pt}
\section{Introduction}
\label{sec:intro}
\vspace{-5pt}

Humans exhibit a remarkable ability to acquire new knowledge continuously from their environments while retaining previously learned information. This capability, known as Continual Learning (CL) \cite{LifeLong_review}, has emerged as a prominent research area within the field of deep learning.
In contrast to traditional deep learning paradigms \cite{DeepRes}, which involve retraining on entire datasets multiple times, CL focuses on developing models that can sustain high performance across all prior tasks while assimilating new information. 
Still, contemporary learning models often experience considerable performance degradation in CL scenarios, primarily due to a phenomenon known as catastrophic forgetting \cite{LifeLong_review}.

The challenge of catastrophic forgetting arises when a model, while learning new tasks, gradually overwrites previously acquired knowledge \cite{LifeLong_review}. To mitigate this, existing work generally falls into three categories: rehearsal-based methods that store a small set of past examples in a memory buffer and replay them during training; 
regularisation methods that add penalty terms to the loss function to prevent drastic changes in important parameters; 
and architecture-based methods that expand the network by adding or removing modules for each incoming task. 
However, most rehearsal-based approaches rely on a single memory buffer, which struggles to balance the retention of both recent and distant experiences under limited capacity, leaving the model vulnerable to forgetting.

Additionally, another major challenge in CL is that most existing methodologies typically assume a static learning setting in which all past data can be revisited repeatedly and each class or task is perfectly balanced. These assumptions are rarely met in real-world scenarios. This focus overlooks a more complex and realistic CL context, characterised by the condition that each data batch is encountered only once, which we refer to as Online Continual Batch-to-Batch Learning (OCBBL). In addition, the data samples in the real-world scenario would be imbalanced, and such a scenario is less explored in OCCBBL. In this paper, we endeavour to tackle catastrophic forgetting under both the OCBBL and the dataset imbalance settings by introducing an innovative memory-based strategy.

The biological findings indicate that knowledge is continuously processed in the brain through both fast and slow memory mechanisms \cite{HumanLerningSystem}. Drawing inspiration from these findings, we introduce an innovative memory framework termed Online Dynamic Expandable Dual Memory (\model). This framework comprises a short-term memory system designed to accommodate dynamic and transient information, alongside a long-term memory system intended to retain enduring knowledge. Specifically, we implement the short-term memory system utilising a buffer, which facilitates the dynamic storage of new data samples while minimising computational overhead. For the long-term memory system, we propose segmenting it into a series of cluster prototypes, each capable of forming an expandable sub-memory buffer to retain data samples that exhibit similar semantic characteristics as seeing more samples from different categories. To tackle the issue of imbalanced data samples, we present a novel sample selection strategy that ensures an equal number of cluster prototypes per category through a $K$-means clustering algorithm. Each cluster prototype serves as the focal point of its corresponding sub-memory buffer, designed to maintain an equivalent number of data samples for each distinct category, thereby ensuring balance across the memory system.

To accumulate critical samples to an appropriate sub-memory buffer, we propose a novel memory optimisation approach that evaluates the knowledge similarity between each cluster prototype and incoming samples via an Optimal Transport (OT) distance. This approach selectively stores the data samples that contain significantly different information from the corresponding cluster prototype. Such a mechanism can promote each sub-memory buffer to store semantically informative data samples, leading to good generalisation performance. Furthermore, we propose a new dynamic memory allocation approach, which automatically assigns more capacity for the short-term memory buffer to provide sufficient training samples in the initial learning phase, which benefits the model convergence. As the number of training steps increases, the proposed approach incrementally assigns more capacities for the long-term memory buffer to preserve more critical past samples, which effectively relieves network forgetting. We also propose a new Divide-and-Conquer (DAC) strategy to reduce the computational cost of the memory optimisation process. Specifically, the DAC strategy recursively groups the incoming samples into a series of cluster prototypes and then combines the samples that are close to each cluster prototype.

We conduct a series of experiments on both standard and imbalanced settings to evaluate the performance of various models, and the empirical results demonstrate that the proposed approach achieves state-of-the-art performance in various CL scenarios. The contributions of this paper are summarised into three parts:
\begin{inparaenum}[(1)]
 \item inspired by the biological findings, we propose a novel training framework that incorporates short- and long-term memory buffers with dynamically changing sizes to address catastrophic forgetting in the OCBBL scenario;
 \item ODEDM, a plug-and-play framework, is designed to be compatible with rehearsal-based models, and it is shown experimentally to enhance their performance;
 \item we also propose a novel DAC approach that can significantly reduce the computational cost of ODEDM.
\end{inparaenum}

\vspace{-1.5em}
\section{Related work}
\label{sec:related_work}

\noindent
Recent advances in OCL have explored memory management and sample selection to mitigate catastrophic forgetting. OnPro \cite{wei2023online} maintains class-wise prototypes for contrastive learning but assumes clean streams and may struggle with data imbalance. OCM \cite{pmlr-v162-guo22g} uses mutual information (MI) maximisation to learn holistic feature representations and preserve past knowledge. It enhances representation robustness via MI objectives between inputs, outputs, and past features, using data augmentations to tighten MI bounds. PuriDivER \cite{bang2022online} handles noisy and ambiguous streams via a hybrid memory strategy and semi-supervised learning on memory partitions. VR-MCL \cite{ICLR2024_0b6df1a9} improves OCL via momentum-based variance reduction for stable Hessian approximation, while POCL \cite{10.5555/3692070.3694280} employs Pareto-optimised gradient aggregation to model inter-task relations and balance past–current performance.

Dual memory systems have been widely adopted to mitigate catastrophic forgetting in continual learning. DGM \cite{kamra2017deep} proposes a generative dual-memory network that consolidates short-term learning via generative replay. DualMEM \cite{MA2023174} emulates the PFC-HC circuitry with a stratified short-term memory and long-term memory design, achieving efficient rehearsal under experience-once constraints. IL2M \cite{9009019} combines exemplar memory with initial class statistics to rectify prediction bias, showing that distillation may not always be beneficial.

Rehearsal-based methods mitigate forgetting by reusing past data and include experience replay, generative replay, and feature replay \cite{10444954}. Experience replay stores past samples for replay using strategies like reservoir sampling \cite{chaudhry2019tinyepisodicmemoriescontinual, riemer2019learninglearnforgettingmaximizing}. DER and DER\texttt{++} \cite{buzzega2020dark} store both samples and logits, while DER\texttt{++}refresh \cite{wang2024unifiedgeneralframeworkcontinual} introduces an ``unlearn-relearn" mechanism to improve adaptability. iCaRL \cite{rebuffi2017icarlincrementalclassifierrepresentation} combines exemplar rehearsal, knowledge distillation \cite{Gou_2021}, and a nearest-mean-of-exemplars rule. Generative replay synthesises past data via auxiliary models, as in DGR \cite{shin2017continuallearningdeepgenerative} and MeRGAN \cite{wu2019memoryreplayganslearning}. Feature replay models past representations, used in GFR \cite{liu2020generativefeaturereplayclassincremental}, FA \cite{iscen2020memoryefficientincrementallearningfeature}, and DSR \cite{zhu2022selfsustainingrepresentationexpansionnonexemplar}. PuriDivER follows this paradigm, using a hybrid memory to balance diversity and purity. Rehearsal-based methods can be combined with the regularization-based technologies to further improve the performance \cite{dhar2019learningmemorizing, iscen2020memoryefficientincrementallearningfeature, Li2016LearningWF, castro2018endtoendincrementallearning, douillard2020podnetpooledoutputsdistillation, Hou2019LearningAU, wu2019memoryreplayganslearning, zhai2019lifelonggancontinuallearning}

\section{Methodology}
\label{sec:proposed_method}


\subsection{Problem Formulation} 

\begin{figure*}[t]
    \centering
\includegraphics[width=\linewidth]{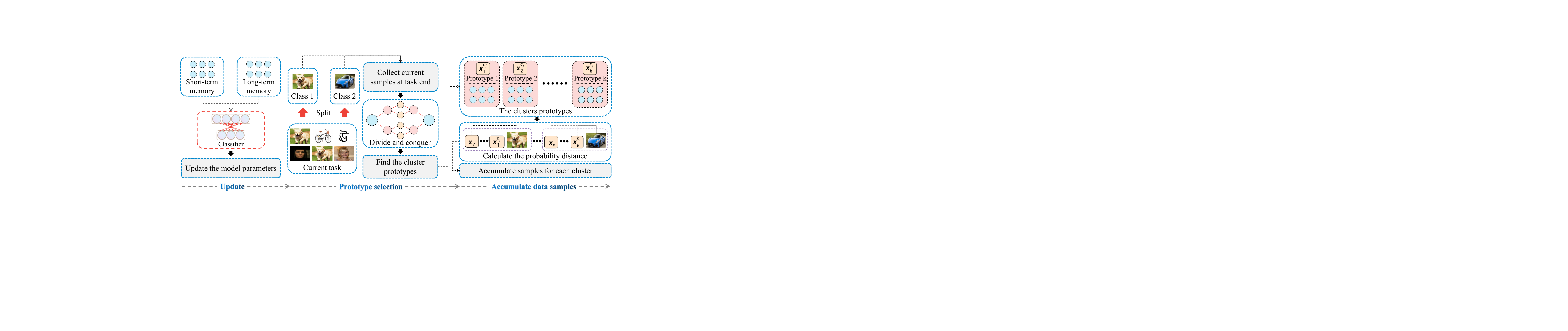}
\vspace{-20pt}
    \caption{The proposed ODEDM approach consists of three steps. In the first step, we employ all stored samples and the incoming data to update the model. The second step first determines several memory clusters using the $K$-means algorithm and then uses the divide-and-conquer method to refine the cluster prototypes. The third step selectively accumulates diverse samples into both memory buffers.}
    \label{fig:dODEDM}
    \vspace{-10pt}
\end{figure*}

In this paper, we focus on this challenging CL scenario where each data batch is only seen once during the training. Let $T = \{T_1, T_2, \dots, T_n\}$ denote a series of $n$ tasks, and each task ${T}_i$ contains a training dataset ${D}_i = \{ {\bf x}_j,{\bf y}_j \}^{n_i}_{j=1}$, where ${\bf x}_j$ and ${\bf y}_j$ represent the sample and the associated label over the data space ${\mathcal{X}} \in {\mathbb{R}}^d$ and the label space ${\mathcal{Y}} \in {\mathbb{R}}^{d_y}$, in the given order. $d$ and  $d_y$ denote the dimensions of the data and label space, respectively. $n_i$ is the total number of data samples from ${D}_i$. In a class-incremental learning scenario, we can create a data stream $S$ by combining these datasets, expressed as $S = \{ D_1,\cdots,D_{n_i}\}$.

Under the OCBBL setting, the data stream $S$ is partitioned into $n'$ sequential data batches, denoted as $\{ ({\bf X}_1, {\bf Y}_1 ),\cdots,({\bf X}_{n'}, {\bf Y}_{n'} )\}$, where each data batch consists of $b$ number of samples. At a certain training time $t_i$ (the $i$-th training time), we can only obtain the associated data batch $({\bf X}_i,{\bf Y}_i)$ and can not access all previous data batches $\{ ({\bf X}_1,{\bf Y}_1),\cdots,({\bf X}_{i-1},{\bf Y}_{i-1})  \}$. A model updated at the $i$-th training time is to minimise the loss across all data batches, expressed as~:
\begin{equation}
    \begin{aligned}
  \theta^\star  =
\underset{ \theta \in \Theta }{\operatorname{argmin}} 
\sum\nolimits^{i}_{j=1} 
\sum\nolimits^{|{\bf X}_j|}_{c=1}
\mathcal{L}_{\rm CE}({\bf y}_{(c,j)}, f_\theta ({\bf x}_{(c,j)}))\,,
\label{optimizationObj}
    \end{aligned}
\end{equation}
where $\Theta$ represents parameter space and $\theta^\star$ is the optimal parameter set that minimises the cross-entropy loss $\mathcal{L}_{\rm CE}$ \cite{cover1999elements} on all previously seen data batches $\{ ({\bf X}_1, {\bf Y}_1 ),\cdots,({\bf X}_{n'}, {\bf Y}_{n'} )\}$. ${\bf x}_{(c,j)}$ and 
${\bf y}_{(c,j)}$ denote the $c$-th labelled sample from the $j$-th data batch $({\bf X}_j,{\bf Y}_j)$. However, optimising \cref{optimizationObj} is intractable in online CL because we can not utilise all previous data batches. To solve this issue, this paper aims to develop a novel memory approach that stores a few past critical samples to minimise the loss defined in \cref{optimizationObj}. 


\subsection{The Memory Structure}
The proposed ODEDM approach consists of a short-term memory system ${\mathcal{M}}^{\rm short}_i$ and a long-term memory system ${\mathcal{M}}^{\rm long}_i$ to preserve both temporary and permanent information, respectively, where the subscript $i$ denotes the memory buffer updated at the $i$-th task learning. For the short-term memory buffer, despite its name, short-term memory retains samples from the entire stream via reservoir sampling, which is computationally efficient. However, the probability of storing new samples decreases over time. This probability should further decrease as memory capacity shrinks for the long-term memory buffer.

For the long-term memory buffer, we aim to preserve more critical and diverse data samples across all previously learned categories. In order to ensure balanced samples in the memory buffer, we propose to define several memory prototypes per category, while each prototype aims to accumulate similar data samples and form a sub-memory buffer. Let $c_l$ denote the $l$-th category and ${\bf x}^{c_l}_h$ represent the $h$-th cluster prototype for the category ${c_l}$. Let ${\mathcal{M}}({c_l},h)$ denote the sub-memory buffer formed using the cluster prototype ${\bf x}^{c_l}_h$. Each sub-memory buffer ${\mathcal{M}}({c_l},h)$ has a maximum memory size, denoted by $\lambda_{max}$.

\subsection{The Sample Selection}
In order to ensure balanced samples in the memory system, we introduce a novel sample selection approach that determines an equal number of cluster prototypes for each category. Specifically, at the end of each task ${T}_i$, we first calculate the number of cluster prototypes required for the category $c_l$ using the function $F_{\rm k}( \rho )$, expressed as~:
\begin{equation}
    \begin{aligned}
        F_{k}( \rho ) = \Big[ \frac{ \rho \lambda_{\rm max} }{n - 1} \cdot \frac{1}{ |{\bf C}_i|}  \Big] \,,
        \label{eq:fk}
    \end{aligned}
\end{equation}
\noindent where ${\bf C}^i$ denotes a set of all categories for $T_i$ using $F_c(D_i)$ and $|{\bf C}_i|$ represents the total number of categories. $\rho$ is a hyperparameter used as the proportion of the long-term memory buffer in the whole memory system. Specifically, we employ the $K$-means algorithm $F_{\rm centre}(D_i(c_l),k)$ to determine the cluster prototypes $\{ {\bf x}^{c_l}_{1},\cdots,{\bf x}^{c_l}_{k} \}$ for the category $c_l$ from the dataset of the $i$-th task learning, where $k = F_k( \rho)$ and $D_i(c_l)$ denotes the data samples from the $j$-th category.

\subsection{Memory Optimisation via Sinkhorn Distance}
To encourage each memory prototype to accumulate more appropriate data samples into the associated sub-memory buffer, we propose to employ a probability distance to evaluate the knowledge similarity between each sub-memory buffer and candidate samples. Specifically, we consider employing the Wasserstein distance \cite{TheWassersteinDistances}, which is a popular probability measure. The main idea of the Wasserstein distance is to quantify the minimum cost of a probability measure $\mathbb{A}$ into another probability measure~$\mathbb{B}$, aiming to solve OT Problems \cite{shen2018wassersteindistanceguidedrepresentation}. Let $(M, \pi)$ be the metric space, where $\pi({\bf x}^1, {\bf x}^2)$ represents the distance between ${\bf x}^1$ and ${\bf x}^2$ in the set $M$. The $p$-th Wasserstein distance is defined as~:
\begin{equation}
    \begin{aligned}
   & W_p(\mathbb{A}, \mathbb{B}) = \left( \inf_{\mu \in \Gamma(\mathbb{A}, \mathbb{B})} \int \rho({\bf x}^1, {\bf x}^2)^p d\mu({\bf x}^1, {\bf x}^2) \right)^{1/p}, 
    \end{aligned}
\end{equation}
\noindent where $\Gamma(\mathbb{A}, \mathbb{B})$ is the set of all joint distributions $\mu({\bf x}^1, {\bf x}^2)$ on $M \times M$ with marginals $\mathbb{A}$ and $\mathbb{B}$. $\mathbb{A} \in \left\{\mathbb{A} : \int \pi({\bf x}^1, {\bf x}^2)^p d\mathbb{A}({\bf x}^1) < \infty, \forall {\bf x}^2 \in M \right\}$.

One of the primary weaknesses of using the Wasserstein distance is its considerable computational costs, which are not suitable when considering using it in our approach. To address this issue, we consider employing the Sinkhorn distance \cite{cuturi2013sinkhorndistanceslightspeedcomputation}, which is one of the OT measures. Specifically, the Sinkhorn distance introduces a smoothing term to transform the optimisation into a strictly convex problem. As a result, this problem can be solved using Sinkhorn's matrix scaling algorithm, which converges rapidly through iterative matrix-vector multiplications, which is computationally efficient. According to \cite{cuturi2013sinkhorndistanceslightspeedcomputation}, we can define the Sinkhorn distance as~:
\begin{equation}
\begin{aligned}
f_{\alpha}({\bf a},{\bf b}) := \min_{P \in U_\alpha( {\bf a},{\bf b})} \left \langle P, M \right \rangle
\,, 
\label{optimizationProblem2}
\end{aligned}
\vspace{-0.5em}
\end{equation} 
where $U_\alpha( {\bf a}, {\bf b} )$ is the convex set, expressed as~:
\begin{align}
    U_\alpha( {\bf a}, {\bf b} ) &:= \{ P \in U({\bf a}, {\bf b}) \,|\, D_{KL}(P \,||\, {\bf a}{\bf b}^{\rm T} ) \le \alpha \} \notag \\
    &\ = \{ P \in U({\bf a}, {\bf b}) \,|\, h'(P) \ge h'({\bf a}) + h'({\bf b}) - \alpha \} \notag \\
    &\  \subset U ({\bf a}, {\bf b})
\,, 
\end{align}
where $D_{KL}(P \,||\, {\bf a}{\bf b}^T) = h'({\bf b}) + h'({\bf a}) - h'(P)$ represents the Kullback-Leibler (KL) divergence~\cite{10.1214/aoms/1177729694}. $U({\bf a}, {\bf b})$ is the transport polytope of ${\bf a}$ and ${\bf b}$, expressed as $U({\bf a}, {\bf b}) := \{ P \in {\mathbb R}^{d \times d}_+ , P {\bf I}_d = {\bf a}, P^{\rm T} {\bf I}_d = {\bf b} \, \}$. $\bf a$ and $\bf b$ are two probability vectors and $h'(\cdot)$ is the entropy function. $P$ is a joint probability and $M$ is a cost matrix mapping of $\bf a$ to $\bf b$. The superscript $\rm T$ denotes the transposition of a matrix. We calculate the distance $f_{\alpha } (\cdot,\cdot)$ using the matrix scaling algorithm \cite{SinkhornDistance}, which receives a pair of data samples. Compared to the Wasserstein distance, the Sinkhorn distance has a faster computing speed and can be evaluated on high-dimensional spaces. Based on \cref{optimizationProblem2}, we can employ $f_{\alpha}$ to choose appropriate samples for each sub-memory buffer.

At the end of each task $T_i$, we determine the cluster prototypes $\{ {\bf x}^{c_l}_{1},\cdots,{\bf x}^{c_l}_{k} \}$ for category $c_l$ using the proposed sample selection approach. Then, we propose a novel Sinkhorn distance-based sample selection approach to accumulate samples for each memory centre ${\bf x}^{c_l}_h,h=1,\cdots,k$, expressed as~:
\begin{equation}
    \begin{aligned}
    {{\mathcal{M}}(c_l,h) } = \{ {\bf x}_v \,|\, 
{\bf x}_v \in {D}_i(c_l), v&=1,\cdots, \lambda-1,\ \\f_{\alpha}({\bf x}_v, {\bf x}^{c_l}_h ) &< f_{\alpha}({\bf x}_{v+1}, {\bf x}^{c_l}_h )    \} \,,
\label{SampleSelectionCriterion}
    \end{aligned}
\end{equation}
\noindent where $D_i(c_l)$ is the subset consisting of $v$ samples (excluding the prototype $h$) of the category ${c_l}$ obtained using the function $ F_{\rm class}(\cdot,\cdot)$, expressed as~:
\begin{equation}
    \begin{aligned}
        F_{\rm class}(D_i, c_l) = \{ {\bf x}_v \,|\, {\bf x}_v \in D_i, v &= 1,\cdots,|D_i|, \\ \ f_{\rm true} ( {\bf x}_v ) & = c_l \}\,,
    \end{aligned}
\end{equation}
\noindent where $f_{\rm true}(\cdot)$ is the function that always returns the true label for the given input ${\bf x}_v$. Specifically, we employ $f_{\alpha} (\cdot,\cdot)$ to calculate a pair of samples in order to reduce computational costs. \cref{SampleSelectionCriterion} aims to choose the data samples that are closest to the cluster prototypes, which can help preserve statistically representative information for $c_l$.

\subsection{Dynamic Memory Allocation}
As the number of tasks increases, using a fixed-capacity short-term memory buffer would fail to preserve sufficient training samples for the model during the initial training stage. In this paper, we address this issue by introducing a novel dynamic memory allocation approach that dynamically allocates samples in short- and long-term memory buffers. As a result, the short-term memory can store sufficient training samples at the initial learning phase and gradually transfers its capacities to the long-term memory buffer as the number of tasks increases. 

Let $\lambda_{\rm max}$ denote the maximum memory size for the proposed memory approach. At the first task learning, the short-term memory buffer can use the full memory capacity (the maximum memory size is $\lambda_{\rm max}$) to store data samples while the long-term memory buffer is 
empty. In the second task learning, we have a default hyperparameter $k$, which is used to denote the number of cluster prototypes per category. Then we find the cluster prototypes and use each one of them to store critical data samples into the associated sub-memory buffer.

Once the $i$-th task learning is finished, we propose to employ \cref{SampleSelectionCriterion} to find a set of sub-memory buffers $\{ {{\mathcal{M}}(c_l,1) },\cdots, {{\mathcal{M}}(c_l,k) } \}$ using \cref{SampleSelectionCriterion}. Then, the long-term memory buffer is updated by~:
\begin{equation}
\begin{aligned}
{\mathcal{M}}^{\rm long}_j = {{\hat{\mathcal{M}}}}^{\rm long}_j \bigcup_{}^{}{{\mathcal{M}}(c_l,1) } \cdots \bigcup_{}^{} {{\mathcal{M}}(c_l,k) }\,, 
\end{aligned}
\vspace{-1em}
\end{equation}
where ${\hat{\mathcal{M}}}^{\rm long}_j$ denotes the long-term memory buffer before the updating process. As a result, we randomly remove data samples of the short-term memory buffer ${\mathcal{M}}^{\rm short}_j$ until the total number of memorised samples is equal to the maximum memory size. This dynamic memory allocation strategy facilitates the progressive expansion of the long-term memory buffer while correspondingly reducing the short-term memory buffer, thereby enhancing the retention of critical historical samples over time.

\subsection{Memory Optimisation Algorithm}

In this section, we provide the detailed optimisation procedure of ODEDM in \Cref{fig:dODEDM}, and we provide the pseudocode of the proposed ODEDM approach in \Cref{alg:ODEDM}, respectively. The algorithmic description of ODEDM consists of three steps~: \\
\noindent
\textbf{Step 1 (Update the model).} We update the model using the cross-entropy loss. \\
\textbf{Step 2 (Find the cluster prototype).} When a task $T_i$ ends, we first select memory clusters per category using \cref{eq:fk}, and then use DAC to recursively refine them. \\
\textbf{Step 3 (Accumulate the data samples).} For each memory cluster, we apply the criterion in \cref{SampleSelectionCriterion} to retain representative samples. The resulting sub-memory buffers are then integrated into the long-term memory. To maintain the overall buffer capacity, an equal number of samples are concurrently discarded from the short-term memory.

\subsection{Divide-and-Conquer Strategy}
A computational defect is observed in ODEDM, which takes a significant runtime for a buffer size of 5120. Therefore, we also propose a method called Divide-and-Conquer (DAC) to mitigate this issue. 
DAC is designed to recursively merge clusters until it reaches the required $k$ for ODEDM to train properly in \textbf{Step 2}. This is particularly useful for large datasets where direct computation of pairwise distances can be computationally expensive. \textbf{Note}: the $K$ in DAC is different from the $k$ in ODEDM. 
In \cref{alg:DAC}, DAC operates as follows~: \\
\noindent \textbf{Step 1 (Initial partition).} Split the input point set into $K$ clusters via the $K$-means algorithm, yielding $\{\mathcal{C}_1,\dots,\mathcal{C}_K\}$. \\
\textbf{Step 2 (Pairwise distances).} Compute the full symmetric distance matrix $D\in\mathbb{R}^{K\times K}$ by applying the Sinkhorn distance of the Wasserstein distance to every pair $(\mathcal{C}_i,\mathcal{C}_j)$. \\
\textbf{Step 3 (Find the closest clusters).} 
Enumerate all subset combinations of clusters, denoted as paths $S = {s_1, \dots, s_{2^K - 1 - K}}$. Among these, identify the subset $s^* \in S$ that yields the shortest intra-subset distance while satisfying the minimum merged size constraint $m$. \\
\textbf{Step 4 \& 5 (Merge and recurse).} 
The shortest clusters in path $s^*$ are merged into an aggregated set, the recursion depth is reduced, and DAC is called recursively. The process stops if no valid path exists ($s^* = \varnothing$) or the depth limit is reached, returning the merged clusters as the final set $\mathcal{X}^*$, constrained by $|\mathcal{X}^*| \ge k$ with $k = m$ as required in ODEDM to perform properly. 
\begin{table*}[ht]
\caption{$ACC_T$ and $\overline{ACC}_T$ comparison under the standard and imbalanced settings, averaged across three independent runs. For each model with ODEDM, the best accuracy is reported among $\rho = 25\%, 50\%, 75\%$ long-term memory proportions. ``-" denotes cases where runtime errors are encountered, thereby preventing the reporting of results for the corresponding models.}
\vspace{-5pt}
\resizebox{\textwidth}{!}{%
\begin{tabular}{@{}lccccccccccccc@{}}
\toprule
\multirow{3}{*}{\textbf{Buffer}}            & \multirow{3}{*}{\textbf{Method}}                                      & \multicolumn{4}{c}{\textbf{CIFAR10}}                                                                                     & \multicolumn{4}{c}{\textbf{CIFAR100}}                                                                                    & \multicolumn{4}{c}{\textbf{TINYIMG}}                                                                \\
                                            &                                                                       & \multicolumn{2}{c}{Standard}                     & \multicolumn{2}{c}{Imbalanced}                                        & \multicolumn{2}{c}{Standard}                     & \multicolumn{2}{c}{Imbalanced}                                        & \multicolumn{2}{c}{Standard}                     & \multicolumn{2}{c}{Imbalanced}                   \\ [2pt]
                                            &                                                                       & \textit{$ACC_5$}           & \textit{$\overline{ACC}_5$}   & \textit{$ACC_5$}           & \textit{$\overline{ACC}_5$}                        & \textit{$ACC_{10}$}           & \textit{$\overline{ACC}_{10}$}   & \textit{$ACC_{10}$}           & \textit{$\overline{ACC}_{10}$}                        & \textit{$ACC_{10}$}           & \textit{$\overline{ACC}_{10}$}   & \textit{$ACC_{10}$}           & \textit{$\overline{ACC}_{10}$}   \\ \midrule
\multirow{18}{*}{$200$}                     & DER \cite{buzzega2020dark}                                            & $21.29_{\pm2.64}$          & $42.25_{\pm5.82}$ & $18.69_{\pm3.88}$          & \multicolumn{1}{c|}{$37.85_{\pm4.10}$} & $4.40_{\pm0.98}$          & $12.12_{\pm1.06}$ & $\mathbf{4.01}_{\pm0.23}$ & \multicolumn{1}{c|}{$10.41_{\pm0.44}$} & $\mathbf{4.33}_{\pm1.06}$ & $9.22_{\pm0.52}$ & $\mathbf{4.02}_{\pm0.56}$ & $7.25_{\pm0.22}$ \\
& DER w/ ODEDM                                                          & $\mathbf{29.99}_{\pm3.48}$ & $48.07_{\pm23.18}$ & $\mathbf{22.04}_{\pm5.85}$ & \multicolumn{1}{c|}{$43.82_{\pm25.43}$} & $\mathbf{4.78}_{\pm0.56}$ & $10.18_{\pm7.50}$ & $3.20_{\pm0.71}$          & \multicolumn{1}{c|}{$8.29_{\pm7.39}$} & $3.83_{\pm0.27}$          & $8.90_{\pm6.02}$ & $2.82_{\pm0.45}$          & $6.68_{\pm4.86}$ \\ \cmidrule(l){3-14} 
& DER\texttt{++} \cite{buzzega2020dark}                                 & $29.67_{\pm4.20}$          & $52.12_{\pm5.88}$ & $28.74_{\pm2.10}$          & \multicolumn{1}{c|}{$45.09_{\pm3.61}$} & $5.73_{\pm0.24}$          & $12.19_{\pm0.96}$ & $3.99_{\pm0.48}$          & \multicolumn{1}{c|}{$9.08_{\pm1.49}$} & $\mathbf{4.97}_{\pm0.63}$ & $10.78_{\pm0.60}$ & $\mathbf{4.43}_{\pm0.85}$ & $9.25_{\pm0.36}$ \\
& DER\texttt{++} w/ ODEDM                                               & $\mathbf{42.71}_{\pm1.86}$ & $57.35_{\pm14.51}$ & $\mathbf{40.95}_{\pm1.82}$ & \multicolumn{1}{c|}{$57.19_{\pm18.63}$} & $\mathbf{7.78}_{\pm1.78}$ & $16.15_{\pm8.92}$ & $\mathbf{7.46}_{\pm1.57}$ & \multicolumn{1}{c|}{$12.51_{\pm6.68}$} & $4.65_{\pm1.16}$          & $12.14_{\pm8.30}$ & $4.09_{\pm0.95}$          & $8.90_{\pm5.50}$ \\ \cmidrule(l){3-14} 
& DER\texttt{++}refresh \cite{wang2024unifiedgeneralframeworkcontinual} & $31.91_{\pm4.80}$          & $55.06_{\pm0.76}$ & $30.24_{\pm8.77}$          & \multicolumn{1}{c|}{$51.92_{\pm2.40}$} & $5.04_{\pm0.64}$          & $10.58_{\pm1.21}$ & $5.43_{\pm1.34}$          & \multicolumn{1}{c|}{$9.90_{\pm0.96}$} & $\mathbf{5.74}_{\pm0.53}$ & $11.20_{\pm0.38}$ & $\mathbf{4.28}_{\pm0.68}$ & $9.14_{\pm0.64}$ \\
& DER\texttt{++}refresh w/ ODEDM                                        & $\mathbf{42.47}_{\pm0.38}$ & $57.91_{\pm13.60}$ & $\mathbf{38.41}_{\pm0.88}$ & \multicolumn{1}{c|}{$52.12_{\pm14.12}$} & $\mathbf{8.47}_{\pm1.24}$ & $15.30_{\pm8.44}$ & $\mathbf{5.68}_{\pm1.99}$ & \multicolumn{1}{c|}{$12.51_{\pm7.57}$} & $5.25_{\pm1.09}$          & $12.85_{\pm8.20}$ & $3.29_{\pm0.73}$          & $9.45_{\pm6.32}$ \\ \cmidrule(l){3-14} 
& FDR \cite{benjamin2018measuring}          & $20.31_{\pm3.38}$          & $39.68_{\pm4.42}$ & $14.75_{\pm1.85}$          & \multicolumn{1}{c|}{$36.24_{\pm5.47}$} & $\mathbf{6.09}_{\pm0.98}$ & $12.89_{\pm1.09}$ & $\mathbf{4.25}_{\pm0.97}$ & \multicolumn{1}{c|}{$9.61_{\pm1.79}$} & $\mathbf{4.55}_{\pm0.57}$ & $10.95_{\pm0.57}$ & $\mathbf{3.55}_{\pm0.17}$ & $8.17_{\pm0.46}$ \\
& FDR w/ ODEDM                                                          & $\mathbf{25.17}_{\pm1.39}$ & $43.30_{\pm25.82}$ & $\mathbf{21.14}_{\pm1.72}$ & \multicolumn{1}{c|}{$40.15_{\pm21.32}$} & $4.14_{\pm0.52}$          & $12.43_{\pm9.49}$ & $3.49_{\pm0.55}$          & \multicolumn{1}{c|}{$8.92_{\pm6.42}$} & $4.18_{\pm0.31}$          & $10.94_{\pm8.13}$ & $3.51_{\pm0.50}$          & $8.56_{\pm5.85}$ \\ \cmidrule(l){3-14} 
& iCaRL \cite{rebuffi2017icarlincrementalclassifierrepresentation}      & $35.32_{\pm2.08}$          & $54.18_{\pm1.64}$ & $34.39_{\pm1.53}$          & \multicolumn{1}{c|}{$52.27_{\pm3.19}$} & $\mathbf{8.56}_{\pm0.37}$ & $16.54_{\pm0.09}$ & $\mathbf{7.41}_{\pm0.30}$ & \multicolumn{1}{c|}{$16.28_{\pm0.97}$} & $3.78_{\pm0.19}$ & $8.83_{\pm0.30}$ & $3.61_{\pm0.05}$ & $8.32_{\pm1.02}$ \\
& iCaRL w/ ODEDM                                                        & $\mathbf{37.64}_{\pm1.56}$ & $58.58_{\pm19.40}$ & $\mathbf{35.26}_{\pm0.58}$ & \multicolumn{1}{c|}{$56.22_{\pm18.40}$} & $1.03_{\pm0.02}$          & $13.53_{\pm14.02}$ & $1.17_{\pm0.34}$          & \multicolumn{1}{c|}{$6.28_{\pm1.33}$} & $-$                         & $-$                & $-$                         & $-$                \\ \cmidrule(l){3-14} 
& PuriDivER \cite{bang2022online}                                       & $21.46_{\pm0.95}$          & $43.64_{\pm24.64}$ & $20.81_{\pm2.00}$          & \multicolumn{1}{c|}{$41.65_{\pm23.46}$} & $7.21_{\pm0.21}$ & $15.70_{\pm8.55}$ & $\mathbf{7.26}_{\pm0.34}$ & \multicolumn{1}{c|}{$14.30_{\pm7.56}$} & $-$                         & $-$                & $-$                         & $-$                \\
& PuriDivER w/ ODEDM                                                    & $\mathbf{23.04}_{\pm1.30}$ & $42.92_{\pm24.40}$ & $\mathbf{22.08}_{\pm1.13}$ & \multicolumn{1}{c|}{$41.08_{\pm23.89}$} & $\mathbf{7.67}_{\pm0.23}$          & $14.67_{\pm7.94}$ & $6.83_{\pm0.43}$          & \multicolumn{1}{c|}{$13.32_{\pm6.56}$} & $-$                         & $-$                & $-$                         & $-$                \\ \cmidrule(l){3-14} &
VR-MCL \cite{ICLR2024_0b6df1a9}                                       & $39.64_{\pm1.60}$ & $56.57_{\pm15.94}$ & 
$36.28_{\pm1.24}$          & \multicolumn{1}{c|}{$54.80_{\pm16.76}$} & $12.75_{\pm0.82}$ & $22.34_{\pm9.32}$ &
$10.28_{\pm0.43}$ & \multicolumn{1}{c|}{$20.45_{\pm9.31}$} & 
$6.87_{\pm0.44}$ & $13.82_{\pm6.65}$ & 
$6.45_{\pm0.52}$ & $11.83_{\pm5.62}$ \\
\multicolumn{1}{c}{}                        & 
VR-MCL w/ ODEDM & 
$\mathbf{41.34}_{\pm4.79}$ & $56.31_{\pm17.71}$ & 
$\mathbf{42.51}_{\pm1.37}$ & \multicolumn{1}{c|}{$57.88_{\pm17.49}$} & 
$\mathbf{15.31}_{\pm0.70}$ & $25.47_{\pm10.77}$ & 
$\mathbf{12.84}_{\pm0.40}$ & \multicolumn{1}{c|}{$21.52_{\pm10.00}$} & 
$8.64_{\pm0.33}$ & $16.67_{\pm8.50}$ & 
$7.70_{\pm0.04}$ & $13.28_{\pm5.94}$ \\ \cmidrule(l){3-14} &
POCL \cite{10.5555/3692070.3694280} & 
$35.08_{\pm2.78}$ & $54.59_{\pm16.92}$ & 
$30.87_{\pm0.29}$ & \multicolumn{1}{c|}{$53.63_{\pm19.05}$} & 
$9.15_{\pm0.92}$ & $20.50_{\pm10.87}$ & 
$\mathbf{8.39}_{\pm0.21}$ & \multicolumn{1}{c|}{$17.18_{\pm7.74}$} & 
$\mathbf{5.49}_{\pm0.75}$ & $12.46_{\pm8.18}$ & 
$\mathbf{4.93}_{\pm0.22}$ & $10.94_{\pm5.70}$ \\
\multicolumn{1}{c}{}                        & 
POCL w/ ODEDM & 
$\mathbf{37.03}_{\pm5.19}$ & $53.85_{\pm15.28}$ & 
$\mathbf{36.90}_{\pm4.76}$ & \multicolumn{1}{c|}{$56.29_{\pm17.37}$} & 
$\mathbf{9.16}_{\pm0.34}$ & $20.43_{\pm9.03}$ & 
$8.31_{\pm0.68}$ & \multicolumn{1}{c|}{$17.74_{\pm7.50}$} & 
$4.77_{\pm0.56}$ & $11.33_{\pm6.63}$ & 
$3.41_{\pm0.52}$ & $9.56_{\pm5.00}$ \\ \midrule
\multicolumn{1}{c}{\multirow{19}{*}{$500$}} & 
DER \cite{buzzega2020dark} & $20.56_{\pm2.64}$  & $42.58_{\pm1.71}$ & $17.25_{\pm2.49}$          & \multicolumn{1}{c|}{$40.19_{\pm1.04}$} & $\mathbf{4.49}_{\pm1.21}$ & $12.96_{\pm1.20}$ & 
$\mathbf{3.89}_{\pm0.60}$ & \multicolumn{1}{c|}{$10.98_{\pm0.63}$} & $4.21_{\pm0.48}$          & $9.57_{\pm0.58}$ & 
$3.17_{\pm0.11}$          & $6.78_{\pm0.79}$ \\
\multicolumn{1}{c}{}                        & DER w/ ODEDM                                                          & $\mathbf{28.34}_{\pm4.98}$ & $48.41_{\pm20.53}$ & $\mathbf{21.32}_{\pm5.02}$ & \multicolumn{1}{c|}{$40.49_{\pm21.07}$} & $4.12_{\pm0.81}$          & $10.52_{\pm7.93}$ & $3.84_{\pm0.34}$          & \multicolumn{1}{c|}{$8.29_{\pm6.13}$} & $\mathbf{4.45}_{\pm0.12}$ & $9.38_{\pm6.72}$ & $\mathbf{3.55}_{\pm0.17}$ & $6.60_{\pm4.31}$ \\ \cmidrule(l){3-14} 
\multicolumn{1}{c}{}                        & DER\texttt{++} \cite{buzzega2020dark}                                 & $39.85_{\pm10.25}$          & $56.48_{\pm4.54}$ & $32.94_{\pm1.58}$          & \multicolumn{1}{c|}{$43.73_{\pm1.81}$} & $8.42_{\pm1.01}$          & $16.12_{\pm2.30}$ & $5.27_{\pm2.54}$          & \multicolumn{1}{c|}{$11.67_{\pm0.99}$} & $6.72_{\pm1.00}$          & $14.32_{\pm1.53}$ & $5.44_{\pm0.06}$          & $10.29_{\pm0.72}$ \\
\multicolumn{1}{c}{}                        & DER\texttt{++} w/ ODEDM                                               & $\mathbf{47.31}_{\pm2.62}$ & $56.02_{\pm9.82}$ & $\mathbf{40.14}_{\pm2.46}$ & \multicolumn{1}{c|}{$59.03_{\pm17.23}$} & $\mathbf{13.62}_{\pm2.51}$ & $18.37_{\pm7.61}$ & $\mathbf{10.16}_{\pm0.55}$ & \multicolumn{1}{c|}{$14.27_{\pm6.46}$} & $\mathbf{7.43}_{\pm0.75}$ & $14.34_{\pm6.10}$ & $\mathbf{5.93}_{\pm1.42}$ & $12.49_{\pm6.24}$ \\ \cmidrule(l){3-14} 
\multicolumn{1}{c}{}                        & DER\texttt{++}refresh \cite{wang2024unifiedgeneralframeworkcontinual} & $43.58_{\pm2.29}$          & $56.51_{\pm1.38}$ & $29.73_{\pm1.82}$          & \multicolumn{1}{c|}{$50.86_{\pm4.54}$} & $9.99_{\pm0.70}$          & $18.13_{\pm3.23}$ & $7.56_{\pm0.98}$          & \multicolumn{1}{c|}{$12.35_{\pm0.83}$} & $6.86_{\pm0.98}$          & $13.03_{\pm1.10}$ & $5.92_{\pm0.53}$          & $11.29_{\pm1.58}$ \\
\multicolumn{1}{c}{}                        & DER\texttt{++}refresh w/ ODEDM                                        & $\mathbf{46.17}_{\pm2.37}$ & $60.04_{\pm14.36}$ & $\mathbf{41.53}_{\pm4.15}$ & \multicolumn{1}{c|}{$54.06_{\pm15.56}$} & $\mathbf{11.79}_{\pm1.91}$ & $17.78_{\pm5.93}$ & $\mathbf{10.27}_{\pm1.19}$ & \multicolumn{1}{c|}{$14.16_{\pm6.77}$} & $\mathbf{7.14}_{\pm0.85}$ & $13.10_{\pm6.37}$ & $\mathbf{7.00}_{\pm0.79}$ & $10.77_{\pm6.06}$ \\ \cmidrule(l){3-14} 
\multicolumn{1}{c}{}                        & FDR \cite{benjamin2018measuring}          & $18.23_{\pm4.39}$          & $42.46_{\pm2.16}$ & $17.36_{\pm2.88}$          & \multicolumn{1}{c|}{$40.41_{\pm1.64}$} & $4.80_{\pm0.69}$          & $12.37_{\pm1.42}$ & $3.49_{\pm0.90}$          & \multicolumn{1}{c|}{$8.55_{\pm0.54}$} & $4.22_{\pm0.10}$          & $10.90_{\pm0.98}$ & $\mathbf{4.17}_{\pm0.96}$ & $8.89_{\pm0.48}$ \\
\multicolumn{1}{c}{}                        & FDR w/ ODEDM                                                          & $\mathbf{24.43}_{\pm1.04}$ & $44.45_{\pm25.30}$ & $\mathbf{17.50}_{\pm1.57}$ & \multicolumn{1}{c|}{$32.94_{\pm18.89}$} & $\mathbf{4.96}_{\pm0.84}$ & $12.20_{\pm7.90}$ & $\mathbf{4.71}_{\pm0.66}$ & \multicolumn{1}{c|}{$10.22_{\pm7.42}$} & $\mathbf{4.84}_{\pm0.38}$ & $10.76_{\pm7.45}$ & $3.90_{\pm0.27}$          & $8.48_{\pm5.77}$ \\ \cmidrule(l){3-14} 
\multicolumn{1}{c}{}                        & iCaRL \cite{rebuffi2017icarlincrementalclassifierrepresentation}      & $35.16_{\pm1.27}$          & $51.72_{\pm5.06}$ & $32.74_{\pm0.74}$          & \multicolumn{1}{c|}{$50.77_{\pm3.66}$} & $\mathbf{9.04}_{\pm0.54}$ & $18.37_{\pm1.75}$ & $\mathbf{8.25}_{\pm0.55}$ & \multicolumn{1}{c|}{$16.16_{\pm1.94}$} & $\mathbf{4.19}_{\pm0.24}$ & $9.44_{\pm1.21}$ & $\mathbf{3.89}_{\pm0.12}$ & $8.64_{\pm0.56}$ \\
\multicolumn{1}{c}{}                        & iCaRL w/ ODEDM                                                        & $\mathbf{38.31}_{\pm0.72}$ & $55.74_{\pm17.55}$ & $\mathbf{36.10}_{\pm0.87}$ & \multicolumn{1}{c|}{$55.64_{\pm18.66}$} & $7.35_{\pm0.14}$          & $16.62_{\pm9.84}$ & $7.49_{\pm0.40}$          & \multicolumn{1}{c|}{$15.44_{\pm8.40}$} & $0.48_{\pm0.08}$          & $7.17_{\pm8.14}$ & $0.51_{\pm0.12}$          & $5.91_{\pm6.81}$ \\ \cmidrule(l){3-14} 
\multicolumn{1}{c}{}                        & PuriDivER \cite{bang2022online}                                       & $21.87_{\pm1.76}$          & $42.92_{\pm24.40}$ & $21.05_{\pm1.21}$          & \multicolumn{1}{c|}{$41.77_{\pm23.59}$} & $\mathbf{8.09}_{\pm1.03}$ & $15.91_{\pm9.00}$ & $6.80_{\pm0.36}$ & \multicolumn{1}{c|}{$13.55_{\pm6.95}$} & $-$                         & $-$                & $-$                         & $-$                \\
\multicolumn{1}{c}{}                        & PuriDivER w/ ODEDM                                                    & $\mathbf{23.73}_{\pm1.83}$ & $42.97_{\pm23.19}$ & $\mathbf{22.76}_{\pm1.26}$ & \multicolumn{1}{c|}{$43.03_{\pm23.78}$} & $8.04_{\pm0.39}$          & $15.57_{\pm7.89}$ & $\mathbf{7.61}_{\pm0.39}$          & \multicolumn{1}{c|}{$12.85_{\pm5.68}$} & $-$                         & $-$                & $-$                         & $-$                \\ \cmidrule(l){3-14} &
VR-MCL \cite{ICLR2024_0b6df1a9}                                       & $46.80_{\pm2.85}$ & $63.44_{\pm15.04}$ & 
$41.92_{\pm4.40}$ & \multicolumn{1}{c|}{$59.79_{\pm16.42}$} & $16.32_{\pm0.64}$ & $25.13_{\pm9.30}$ & 
$13.43_{\pm0.78}$ & \multicolumn{1}{c|}{$20.79_{\pm6.38}$} & 
$10.08_{\pm0.40}$ & $15.99_{\pm5.96}$ & 
$9.45_{\pm0.08}$ & $15.56_{\pm5.80}$ \\
\multicolumn{1}{c}{}                        & 
VR-MCL w/ ODEDM & 
$\mathbf{47.50}_{\pm1.93}$ & $63.13_{\pm17.15}$ & 
$\mathbf{47.70}_{\pm3.55}$ & \multicolumn{1}{c|}{$55.32_{\pm9.22}$} & 
$\mathbf{18.32}_{\pm1.07}$ & $27.35_{\pm9.20}$ & 
$\mathbf{16.54}_{\pm0.47}$ & \multicolumn{1}{c|}{$24.09_{\pm8.28}$} & 
$\mathbf{10.74}_{\pm0.59}$ & $18.22_{\pm7.08}$ & 
$\mathbf{10.60}_{\pm0.82}$ & $16.13_{\pm5.41}$ \\ \cmidrule(l){3-14} &
POCL \cite{10.5555/3692070.3694280}                                       & $44.45_{\pm6.02}$ & $57.88_{\pm11.48}$ & 
$42.21_{\pm5.08}$ & \multicolumn{1}{c|}{$61.21_{\pm16.79}$} & $12.75_{\pm0.55}$ & $23.01_{\pm8.89}$ & 
$12.12_{\pm0.59}$ & \multicolumn{1}{c|}{$20.46_{\pm6.84}$} & 
$6.34_{\pm0.37}$ & $13.13_{\pm5.73}$ & 
$\mathbf{6.73}_{\pm0.26}$ & $12.88_{\pm6.10}$ \\
\multicolumn{1}{c}{}                        & 
POCL w/ ODEDM                                                    & 
$\mathbf{48.48}_{\pm0.94}$ & $64.05_{\pm13.00}$ & 
$\mathbf{44.29}_{\pm2.09}$ & \multicolumn{1}{c|}{$60.80_{\pm15.53}$} & 
$\mathbf{15.82}_{\pm1.48}$ & $25.39_{\pm7.32}$ & 
$\mathbf{14.03}_{\pm0.65}$ & \multicolumn{1}{c|}{$23.32_{\pm6.67}$} & 
$\mathbf{6.91}_{\pm0.46}$ & $14.07_{\pm6.38}$ & 
$6.47_{\pm0.27}$ & $11.29_{\pm4.20}$ \\ \midrule
\multirow{17}{*}{$5120$}                    & DER \cite{buzzega2020dark}                                            & $24.65_{\pm2.26}$          & $44.16_{\pm0.52}$ & $17.38_{\pm2.60}$          & \multicolumn{1}{c|}{$37.97_{\pm5.05}$} & $4.26_{\pm0.22}$          & $12.61_{\pm0.74}$ & $\mathbf{4.43}_{\pm0.31}$ & \multicolumn{1}{c|}{$11.04_{\pm0.37}$} & $4.30_{\pm0.26}$          & $9.00_{\pm0.05}$ & $\mathbf{3.81}_{\pm0.28}$ & $7.61_{\pm0.42}$ \\
                                            & DER w/ ODEDM                                                          & $\mathbf{29.96}_{\pm1.63}$ & $46.82_{\pm19.38}$ & $\mathbf{21.89}_{\pm4.60}$ & \multicolumn{1}{c|}{$40.51_{\pm25.19}$} & $\mathbf{4.83}_{\pm0.33}$ & $11.23_{\pm10.10}$ & $2.92_{\pm0.78}$          & \multicolumn{1}{c|}{$8.03_{\pm5.39}$} & $\mathbf{4.49}_{\pm0.48}$ & $9.29_{\pm5.78}$ & $3.55_{\pm0.37}$          & $7.28_{\pm4.40}$ \\ \cmidrule(l){3-14} 
                                            & DER\texttt{++} \cite{buzzega2020dark}                                 & $43.31_{\pm9.12}$          & $53.53_{\pm3.32}$ & $39.16_{\pm3.28}$          & \multicolumn{1}{c|}{$51.66_{\pm2.82}$} & $11.98_{\pm1.84}$          & $17.74_{\pm0.55}$ & $7.89_{\pm3.14}$          & \multicolumn{1}{c|}{$11.88_{\pm1.51}$} & $9.96_{\pm1.80}$          & $15.55_{\pm0.12}$ & $8.05_{\pm0.78}$          & $12.19_{\pm0.88}$ \\
                                            & DER\texttt{++} w/ ODEDM                                               & $\mathbf{54.49}_{\pm1.44}$ & $60.36_{\pm10.89}$ & $\mathbf{47.90}_{\pm2.10}$ & \multicolumn{1}{c|}{$53.82_{\pm11.45}$} & $\mathbf{22.42}_{\pm5.67}$ & $24.84_{\pm6.90}$ & $\mathbf{13.62}_{\pm3.94}$ & \multicolumn{1}{c|}{$17.55_{\pm6.00}$} & $\mathbf{15.61}_{\pm3.29}$ & $20.09_{\pm4.61}$ & $\mathbf{11.13}_{\pm1.72}$ & $15.11_{\pm3.65}$ \\ \cmidrule(l){3-14} 
                                            & DER\texttt{++}refresh \cite{wang2024unifiedgeneralframeworkcontinual} & $47.78_{\pm3.63}$          & $55.67_{\pm5.27}$ & $34.19_{\pm4.12}$          & \multicolumn{1}{c|}{$50.59_{\pm6.00}$} & $8.33_{\pm1.00}$          & $18.59_{\pm1.66}$ & $5.81_{\pm1.90}$          & \multicolumn{1}{c|}{$12.47_{\pm1.55}$} & $10.21_{\pm0.49}$          & $15.91_{\pm1.16}$ & $7.34_{\pm0.24}$          & $11.51_{\pm0.29}$ \\
                                            & DER\texttt{++}refresh w/ ODEDM                                        & $\mathbf{55.17}_{\pm2.30}$ & $62.97_{\pm12.66}$ & $\mathbf{45.22}_{\pm4.63}$ & \multicolumn{1}{c|}{$51.49_{\pm14.41}$} & $\mathbf{21.60}_{\pm3.27}$ & $24.46_{\pm6.93}$ & $\mathbf{15.15}_{\pm1.92}$ & \multicolumn{1}{c|}{$18.60_{\pm8.99}$} & $\mathbf{17.65}_{\pm0.43}$ & $21.97_{\pm5.78}$ & $\mathbf{12.71}_{\pm1.17}$ & $15.49_{\pm3.61}$ \\ \cmidrule(l){3-14} 
                                            & FDR \cite{benjamin2018measuring}          & $\mathbf{19.33}_{\pm2.17}$ & $39.77_{\pm1.59}$ & $13.47_{\pm0.71}$          & \multicolumn{1}{c|}{$36.72_{\pm4.82}$} & $4.97_{\pm0.78}$          & $12.52_{\pm1.32}$ & $4.23_{\pm1.41}$          & \multicolumn{1}{c|}{$9.62_{\pm0.15}$} & $\mathbf{5.06}_{\pm0.38}$ & $11.15_{\pm0.56}$ & $4.20_{\pm0.63}$          & $9.02_{\pm0.34}$ \\
                                            & FDR w/ ODEDM                                                          & $18.33_{\pm0.58}$          & $38.99_{\pm24.61}$ & $\mathbf{17.86}_{\pm0.14}$ & \multicolumn{1}{c|}{$35.77_{\pm22.49}$} & $\mathbf{6.04}_{\pm0.94}$ & $13.84_{\pm10.31}$ & $\mathbf{4.94}_{\pm0.12}$ & \multicolumn{1}{c|}{$11.14_{\pm7.98}$} & $5.01_{\pm0.23}$          & $11.88_{\pm7.48}$ & $\mathbf{4.69}_{\pm0.22}$ & $8.84_{\pm5.40}$ \\ \cmidrule(l){3-14} 
                                            & iCaRL \cite{rebuffi2017icarlincrementalclassifierrepresentation}      & $47.12_{\pm0.70}$          & $57.42_{\pm3.19}$ & $34.48_{\pm1.25}$          & \multicolumn{1}{c|}{$52.33_{\pm0.87}$} & $\mathbf{9.46}_{\pm0.13}$ & $17.56_{\pm0.69}$ & $\mathbf{8.41}_{\pm0.54}$ & \multicolumn{1}{c|}{$15.81_{\pm1.79}$} & $\mathbf{4.09}_{\pm0.31}$ & $9.75_{\pm0.72}$ & $\mathbf{3.97}_{\pm0.18}$ & $8.26_{\pm0.35}$ \\
                                            & iCaRL w/ ODEDM                                                        & $\mathbf{47.50}_{\pm0.88}$ & $64.95_{\pm14.69}$ & $\mathbf{44.78}_{\pm2.11}$ & \multicolumn{1}{c|}{$60.86_{\pm15.84}$} & $9.28_{\pm0.14}$          & $15.95_{\pm7.15}$ & $8.33_{\pm0.16}$          & \multicolumn{1}{c|}{$15.51_{\pm7.90}$} & $4.01_{\pm0.02}$          & $9.61_{\pm6.72}$ & $3.85_{\pm0.18}$          & $8.63_{\pm5.83}$ \\ \cmidrule(l){3-14} &
VR-MCL \cite{ICLR2024_0b6df1a9}                                       & $52.36_{\pm1.78}$ & $63.44_{\pm14.73}$ & 
$44.76_{\pm4.25}$ & \multicolumn{1}{c|}{$54.64_{\pm13.39}$} & $21.89_{\pm0.68}$ & $29.35_{\pm7.28}$ & 
$16.88_{\pm0.94}$ & \multicolumn{1}{c|}{$23.89_{\pm6.07}$} & 
$16.74_{\pm0.92}$ & $22.75_{\pm5.83}$ & 
$13.22_{\pm0.08}$ & $17.35_{\pm4.10}$ \\
\multicolumn{1}{c}{}                        & 
VR-MCL w/ ODEDM                                                    & 
$\mathbf{55.39}_{\pm0.56}$ & $68.03_{\pm13.57}$ & 
$\mathbf{47.70}_{\pm3.23}$ & \multicolumn{1}{c|}{$56.52_{\pm14.14}$} & 
$\mathbf{25.37}_{\pm0.97}$ & $31.19_{\pm6.60}$ & 
$\mathbf{22.41}_{\pm0.99}$ & \multicolumn{1}{c|}{$27.48_{\pm6.17}$} & 
$\mathbf{19.69}_{\pm0.31}$ & $24.18_{\pm5.36}$ & 
$\mathbf{17.49}_{\pm0.78}$ & $19.71_{\pm3.49}$ \\ \cmidrule(l){3-14} &
POCL \cite{10.5555/3692070.3694280}                                       & $61.65_{\pm0.39}$ & $66.60_{\pm10.69}$ & 
$59.07_{\pm1.83}$ & \multicolumn{1}{c|}{$69.30_{\pm10.29}$ } & 
$29.29_{\pm1.09}$ & $33.88_{\pm4.88}$ & 
$26.40_{\pm2.21}$ & \multicolumn{1}{c|}{$30.86_{\pm6.44}$} & 
$19.61_{\pm0.53}$ & $24.09_{\pm4.38}$                & 
$16.64_{\pm0.56}$ & $19.56_{\pm3.96}$ \\
\multicolumn{1}{c}{}                        
& POCL w/ ODEDM                                                    & 
$\mathbf{66.74}_{\pm1.13}$ & $74.38_{\pm7.52}$ & 
$\mathbf{62.01}_{\pm3.15}$ & \multicolumn{1}{c|}{$63.99_{\pm5.81}$} & 
$\mathbf{35.56}_{\pm0.61}$ & $36.34_{\pm4.23}$ & 
$\mathbf{29.56}_{\pm4.06}$ & \multicolumn{1}{c|}{$27.75_{\pm4.91}$} & 
$\mathbf{21.15}_{\pm0.62}$ & $24.26_{\pm3.07}$ & 
$\mathbf{19.58}_{\pm0.24}$ & $22.21_{\pm2.61}$ \\ \bottomrule
\end{tabular}%
}
\label{tab:accResults}
\vspace{-1.5em}
\end{table*}

\vspace{-5pt}
\section{Experiments}


\subsection{Experiment Settings}
Specifically, we consider many standard CL datasets and split each dataset into several tasks, and then we split CIFAR10 \cite{krizhevsky2009learning} and TINYIMG \cite{Wu2017TinyIC} into 5 and 10 tasks according to \cite{Delange_2021, zenke2017continuallearningsynapticintelligence}, 
each of which contains 2 and 20 classes, in the stated order. We also follow \cite{wang2024unifiedgeneralframeworkcontinual} to split CIFAR100 \cite{krizhevsky2009learning} into 10 tasks, each of which contains samples from ten classes.
Also, we consider a more challenging setting, called the imbalanced learning setting, where it takes out the samples indexed by even numbers, such as $\{0, 2, 4, ... \ n\} \in class$ for the proposed dual buffer and dynamic dual buffer methods, resulting in IMB-CIFAR10, IMB-CIFAR100 and IMB-TINYIMG. 


\noindent
\textbf{Baselines.} We augment methods including DER, DER\texttt{++}, DER\texttt{++}refresh, FDR, iCaRL PuriDivER, VR-MCL, and POCL with and without our method. We do not perform buffer fitting for PuriDivER to ensure a fair comparison, as this step is not applied to the other baselines.

\noindent
\textbf{Training.} To ensure a fair comparison, we set the trade-off coefficients to $\alpha = 0.1$ and $\beta = 0.5$ for DER-based methods, and $\alpha = 0.1$ for FDR. We employ the Stochastic Gradient Descent (SGD) \cite{sgd} optimiser and set the batch size to 32 for all model training. 
In this paper, we consider a more challenging setting, OCBBL, where a new data batch can be accessed only once. Thus, the epoch is 1 for all models \cite{10444954}. 
For the proposed ODEDM, we proportionate long-term memory size to 25\%, 50\%, and 75\% of the total buffer size. A conversion table from $k$ to proportion of long-term memory size is shown in \Cref{tab:k_approx} in \Cref{sec:additional_results}. 
We use the framework, mammoth, from \cite{buzzega2020dark, boschini2022class} to conduct the experiments.

\noindent
\textbf{Network architecture.} We use ResNet-18 \cite{DeepRes} as the backbone network for all experiments in \cref{tab:accResults}.

\noindent
\textbf{Evaluation.} Following prior work in CL~\cite{gao2023unifiedcontinuallearningframework,wang2022learningpromptcontinuallearning, liang2024inflorainterferencefreelowrankadaptation}, 
we assess model performance using two standard metrics, namely, the last accuracy $ACC_T$ and the average accuracy 
$\overline{ACC}_T = \frac{1}{T}\sum_{i=1}^{T} ACC_i$, 
where $T$ is the total number of tasks. 
Each $ACC_i$ quantifies the mean performance after training on the $i$-th task, defined as~:
\begin{equation}
ACC_i = \frac{1}{i}\sum\nolimits_{j=1}^{i} a_{i,j},
\end{equation}
where $a_{i,j}$ denotes the accuracy on task $j$ once the model completes learning task $i$. The \textit{forgetting rate} (FR) quantifies the performance decline on previously learned tasks as new tasks are introduced, reflecting the severity of catastrophic forgetting. It is defined as
\begin{equation}
\text{FR}_i = 100\% - ACC_i,
\end{equation}
where $ACC_i$ denotes the last accuracy after learning $i$ tasks.

\noindent
\textbf{DAC parameters.} DAC has two hyperparameters, $K$ and $depth$, set in ranges $K\in[3,5]$ and $depth\in[2,4]$ for each long-term memory proportions $\rho = 25\%, 50\%, 75\%$. We tune them using ODEDM on CIFAR10 with DER\texttt{++}refresh. Hyperparameter analysis is detailed in \Cref{tab:proportion} in \Cref{sec:additional_results}. The optimal values are: for buffer sizes 200 and 500 $(K=3,\ depth=4)$ at 25\%, $(K=3,\ depth=2)$ at 50\%, and $(K=5,\ depth=3)$ at 75\%; for buffer size 5120 $(K=5,\ depth=3)$ across all settings for efficiency.

\noindent
\textbf{Ablation.} DER\texttt{++}refresh on CIFAR10 with a buffer size of 200 is evaluated under three configurations: (1) mixup, which averages samples within each cluster to form prototypes and serves as the vanilla baseline of ODEDM; (2) ODEDM without DAC; and (3) ODEDM with DAC. Also, we compare different distance metrics used in ODEDM, including L1, L2, Maximum Mean Discrepancy (MMD) with the RBF kernel \cite{DBLP:journals/corr/abs-0805-2368,10.1145/130385.130401}, and Sinkhorn distances.

\begin{figure*}[ht]
  \centering
  \begin{minipage}{\textwidth}
    \centering
    \includegraphics[width=\linewidth]{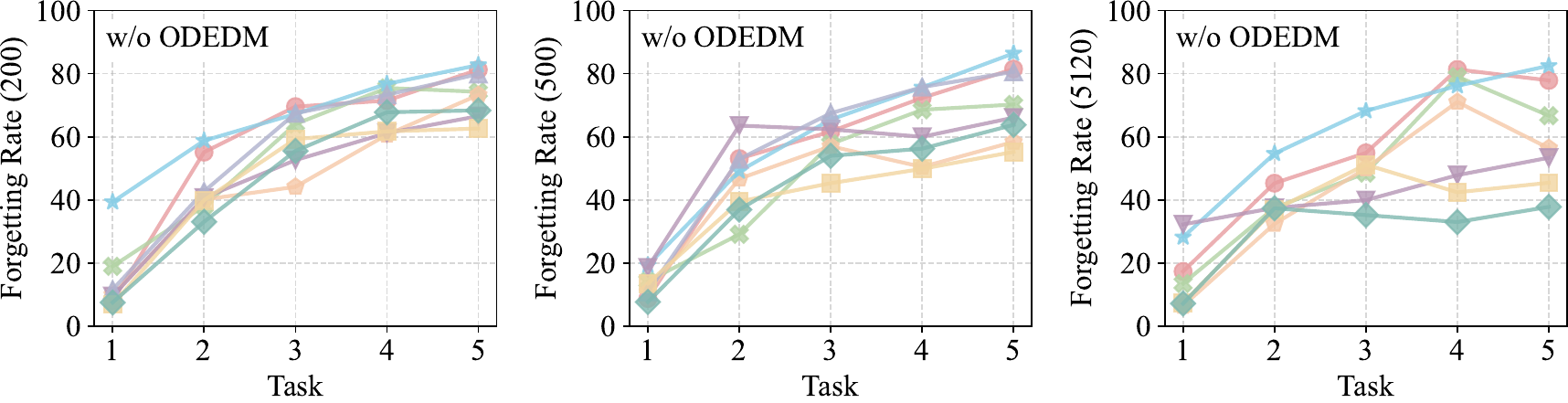}
  \end{minipage}
  \begin{minipage}{\textwidth}
      \centering
      \includegraphics[width=\linewidth]{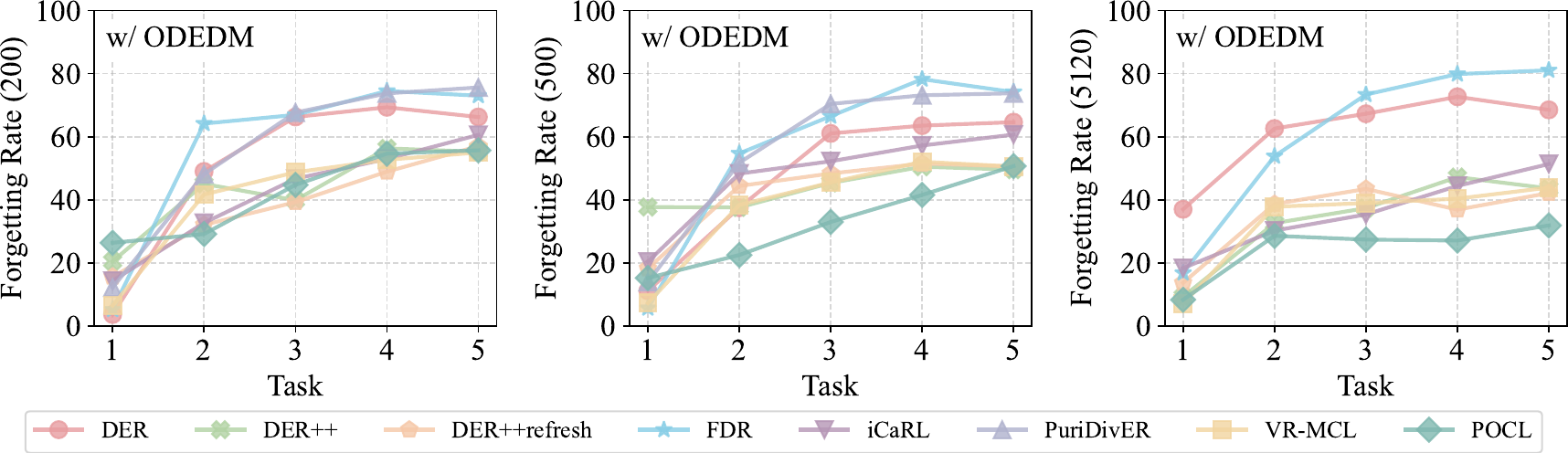}
  \end{minipage}
\caption{Forgetting rate analysis on CIFAR10, obtained from various models trained with different memory configurations. The results report forgetting scores ($\downarrow$) for buffer sizes of 200, 500, and 5120 under the standard setting.}
\label{fig:forgetting_curve}
\vspace{-1.5em}
\end{figure*}

\subsection{Empirical Results}

\textbf{Standard setting.} The $ACC_T$ and $\overline{ACC}_T$ of models employing ODEDM, as reported in \Cref{tab:accResults}, corresponds to the optimal accuracy presented in \Cref{tab:cifar10_full,tab:cifar100_full,tab:tinyimg_full} in \Cref{sec:additional_results}, given different proportions of long-term memory size. Augmenting rehearsal-based methods with the proposed ODEDM substantially enhances the $ACC_T$ and $\overline{ACC}_T$ on CIFAR10 across all buffer sizes. However, its effectiveness on CIFAR100 and TINYIMG is occasionally limited, likely due to the increased complexity of the two datasets. DER\texttt{++}, DER\texttt{++}refresh, VR-MCL, and POCL typically achieve notable performance gains when augmented with ODEDM.
For example, DER\texttt{++} on CIFAR10 with a buffer of size $200$ increases from 29.7\% to 42.7\% when using ODEDM. Meanwhile, the largest buffer size of $5120$, DER\texttt{++} with ODEDM, raises $ACC_5$ performance from 43.3\% to 54.5\%. 
Similar trends hold in CIFAR100 and TINYIMG, especially when DER\texttt{++}, DER\texttt{++}refresh, VR-MCL, and POCL are augmented with ODEDM, leading to substantial relative performance gains of around 80\%–170\% for a buffer size of 5120. In contrast, $ACC_T^{\text{Task-IL}}$ defined in \cref{tab:proportion} from \Cref{sec:additional_results}, which assumes known task identities, sees slight decreases, reflecting that our model primarily enhances cross-task representation rather than within-task discrimination. 

Rehearsal-based methods vary in their compatibility with ODEDM. DER\texttt{++}refresh gains more from ODEDM than DER\texttt{++} and DER, widening the performance gap, as its refresh mechanism captures more benefits of ODEDM, leading to smaller yet meaningful improvements. FDR, despite its lower baseline accuracy, achieves gains of 4–5 percentage points on CIFAR10 and CIFAR100 when augmented with ODEDM. Similarly, PuriDivER exhibits marginal improvements when not fitting the buffer at the end of the task. Conversely, iCaRL collapses on TINYIMG under a buffer size of 200, which may be attributed to incompatibilities between its nearest-mean classifier and the feature representations induced by ODEDM. Its performance shows limited variability when integrated with ODEDM, occasionally exhibiting degradation. 

In \cref{tab:cifar10_taskil_only,tab:cifar100_taskil_only,tab:tinyimg_taskil_only} from \Cref{sec:additional_results}, across most of the experiments, $ACC_T^{\text{Task-IL}}$ results exceed $ACC_T$. ODEDM excels at mitigating inter-task interference in $ACC_T$, rather than enhancing intra-task discrimination.


\begin{figure*}[t]
  \centering
  \begin{subfigure}[b]{0.32\linewidth}
    \centering
    \includegraphics[width=\linewidth]{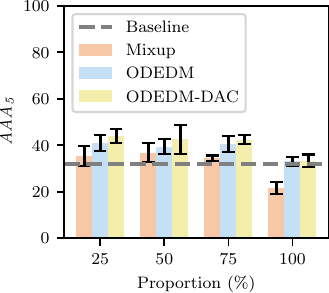}
    \caption{Ablation results.}
    \label{fig:ablation}
  \end{subfigure}\hfill
  \begin{subfigure}[b]{0.32\linewidth}
    \centering
    \includegraphics[width=\linewidth]{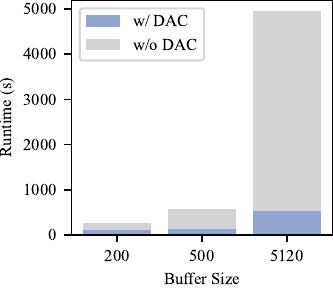}
    \caption{DAC runtime.}
    \label{fig:DAC_time}
  \end{subfigure}\hfill
  \begin{subfigure}[b]{0.32\linewidth}
    \centering
    \includegraphics[width=\linewidth]{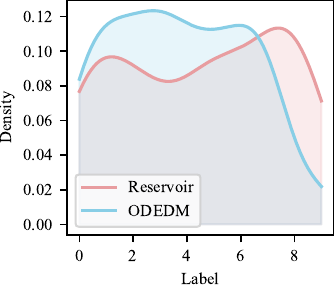}
    \caption{Buffer distribution.}
    \label{fig:buffer_distribution}
  \end{subfigure}
  \caption{(a) The ablation study of the proposed ODEDM with mixup, DAC and without DAC. (b) The runtime difference between with DAC and without DAC. (c) The buffer distribution of the proposed ODEDM with DAC on CIFAR10 under the imbalanced setting.}
  \vspace{-1.6em}
\end{figure*}

\noindent
\textbf{Imbalanced setting.} In \Cref{tab:accResults}, under imbalanced class distributions, all methods exhibit a decline in performance compared to the standard setting. However, the integration of our model mitigates this degradation, particularly for the DER family, VR-MCL and POCL. For instance, on CIFAR10 with a buffer size of 200, DER\texttt{++} augmented with ODEDM achieves 40.9\% $ACC_5$ accuracy under imbalance, compared to 28.7\% without ODEDM. Similarly, on CIFAR100 with a buffer size of 5120, DER\texttt{++}refresh with ODEDM improves from 5.8\% to 15.2\%. In contrast, for TINYIMG, where class imbalance induces a severe performance drop, ODEDM yields only marginal improvements over its non-augmented counterparts, showing a greater challenge in this dataset.

\noindent
\textbf{Forgetting rate.} 
To assess the effectiveness of the proposed method in mitigating catastrophic forgetting across sequential tasks, we perform a forgetting rate analysis on CIFAR10 under the standard setting, considering buffer sizes of 200, 500, and 5120. At each task transition, the classification error on all previously encountered tasks is computed, and the results are presented in \cref{fig:forgetting_curve}. The empirical results demonstrate that models augmented with ODEDM consistently achieve lower forgetting scores (smaller score is better) across all tasks compared to their non-augmented baselines. The findings indicate that ODEDM enhances the preservation of informative samples, thereby improving the model’s ability to counteract network forgetting over time.

\noindent
\textbf{Short- and long-term memory.} 
To examine the interplay between short- and long-term memory in ODEDM, we conduct a hyperparameter analysis by varying the proportion allocated to long-term memory (25\%, 50\%, and 75\%) on CIFAR10, CIFAR100, and TINYIMG. The experimental configurations are summarised in \cref{tab:k_approx} in \Cref{sec:additional_results}, with detailed results reported in \cref{tab:cifar10_full,tab:cifar100_full,tab:tinyimg_full} in \Cref{sec:additional_results}. Several key observations emerge: (1) allocating 25\% or 50\% of memory to the long-term buffer generally yields improved performance across datasets; (2) classification accuracy is sensitive to the proportion between short- and long-term memory; but (3) when the memory capacity is large, performance becomes relatively insensitive.
\noindent
\textbf{Buffer distribution.} 
We evaluate memory optimisation by training DER\texttt{++}refresh with 75\% long-term memory on CIFAR10 using ODEDM. After training, we record the number of samples per class and use Kernel Density Estimation (KDE) to visualise the buffer distribution in \cref{fig:buffer_distribution}. The results show our approach maintains balanced samples across earlier tasks while reducing samples for later tasks, preserving essential knowledge. In contrast, reservoir sampling yields a multimodal distribution, highlighting the robustness of our method under the OCBBL setting.

\noindent
\textbf{Large-scale dataset.} \cref{tab:imgnet-r} (in \Cref{sec:additional_results}) shows that $ACC_{10}$ remains low ($\approx$1.0\%) due to ImageNet-R \cite{hendrycks2021many}’s complexity, yet ODEDM still exceeds methods without it.

\noindent
\textbf{Other OCL baselines.} 
In \cref{tab:onpro} (see \Cref{sec:additional_results}), under the online setting, A-GEM \cite{chaudhry2018efficient} and GSS \cite{aljundi2019gradient} perform worst, while ER \cite{chaudhry2019tiny}, MIR \cite{aljundi2019online}, and GDumb \cite{prabhu2020gdumb} show moderate gains (up to 34.0\% at $M=500$). CoPE \cite{de2021continual} and DER\texttt{++} achieve 42.9\% and 39.9\%, respectively. Offline training (50 epochs) yields minor improvements, with FDR performing best among baselines. DER\texttt{++} with ODEDM attains the highest accuracy (42.7\% at $M=200$, 47.3\% at $M=500$), surpassing offline models.


\vspace{-2.5pt}
\subsection{Ablation Study}
\vspace{-2.5pt}
\textbf{The DAC configuration.} We use DER\texttt{++}refresh as the baseline method. In \cref{fig:ablation}, we can observe that ODEDM with DAC achieves the best performance. The vanilla method (mixup) achieves slightly better performance than the baseline, and setting $\rho$ to 100\% hinders the performance. This indicates that ODEDM with DAC not only improves the performance of ODEDM but also significantly speeds up the training process, as shown in \cref{fig:DAC_time}, measured by a NVIDIA GeForce RTX 3090.

\noindent
\textbf{Distance metric analysis.} We use Principal Component Analysis (PCA) \cite{jolliffe2002principal} to visualise the distribution of long-term memory samples in CIFAR10. To quantify how well each distance metric preserves the underlying data manifold, we map the long-term memory buffer to partial dataset embeddings via ordinary least squares and compute cosine similarities. While L2 and Sinkhorn show the highest alignment (\cref{fig:pca}) and all metrics yield comparable similarities, their performance varies significantly. For DER\texttt{++}refresh with ODEDM, L1 attains an $ACC_5$ accuracy of 41.09\%, L2 achieves 44.00\%, Maximum Mean Discrepancy (MMD) \cite{10.5555/2188385.2188410} reaches 36.93\%, and the Sinkhorn obtains 45.01\%. This indicates that metric choice influences replay effectiveness even when similarity scores are close. Sample selections for each metric are visualised in \cref{fig:l1_samples,fig:l2_samples,fig:mmd_samples,fig:sinkhorn_samples} in \Cref{sec:additional_results}.



\vspace{-8pt}
\section{Conclusion and Limitations}
\vspace{-5pt}
This paper tackles catastrophic forgetting in OCL by introducing ODEDM, a dynamic memory allocation method that integrates short- and long-term buffers to preserve critical samples. Allocating 25\% or 50\% to short-term memory yields better performance. A lightweight strategy, DAC, further improves efficiency. Experiments on CIFAR10 show consistent gains for rehearsal-based models, especially for DER variants and recent models such as VR-MCL and POCL. Moreover, DER\texttt{++} with ODEDM outperforms other OCL baselines and even some 50-epoch models. The main limitation is scalability to larger datasets such as CIFAR100 and TINYIMG, which we leave for future work.


{
    \small
    \bibliographystyle{ieeenat_fullname}
    \bibliography{main}
}

\newpage
\clearpage
\setcounter{page}{1}

\appendix

\section*{Appendix}
\section{Algorithmic Details}
\label{sec:algo_details}

In this section, we detail the algorithmic formulation of the proposed Online Dynamic Expandable Dual Memory (ODEDM) framework, enhanced with a Divide-and-Conquer (DAC) strategy designed to alleviate computational burden.

\subsection{Online Dynamic Expandable Dual Memory}
As illustrated in \Cref{alg:ODEDM}, ODEDM is invoked at the end of each task. 
The procedure begins by identifying all classes present in the samples observed during the current task. 
For each class, the samples are partitioned into $k$ clusters, from which a set of representative prototypes is derived.
Next, the distance between every sample and each prototype within a class is computed, and for each prototype, the nearest sample is selected. 
These selected samples are subsequently committed to the long-term memory buffer. 
To preserve a fixed memory budget, the short-term buffer is then updated by randomly discarding an equal number of samples to those newly added to the long-term buffer. 
This ensures balanced memory allocation while maintaining the representativeness of stored samples.

\subsection{Divide-and-Conquer}
As presented in \Cref{alg:DAC}, the DAC module is designed to extract a representative subset of samples from all observations within the current task. 
Its interaction with ODEDM follows a two-stage process. First, DAC compresses the full set of task samples by selecting a subset that preserves the overall distributional characteristics of the original data. 
ODEDM then operates on this reduced subset to determine which samples should be retained in the long-term memory buffer. 
For example, in a task containing 10,000 samples, DAC may reduce the candidate set to roughly 3,000 samples, depending on its hyperparameter configuration.
Operationally, DAC begins by partitioning all samples into $K$ initial clusters. 
It then enumerates all feasible traversal paths across these clusters and identifies the path with the minimum accumulated inter-cluster distance. 
Clusters along this path are iteratively merged until the number of clusters is reduced to the target $k$ required by ODEDM. 
This process ensures that the resulting subset remains compact while retaining the structural diversity necessary for effective long-term memory selection.


\begin{algorithm}[t]
\caption{The training of ODEDM}
\label{alg:ODEDM}
\textbf{Input:} The total number of tasks $n$ \\
the proportion $\rho$ of ${\mathcal{M}}^{\rm long}_i$, \\
the model's initial parameters\\
\textbf{Output:} Selected samples in long-term memory\\
\For{$i = 1$ \textbf{to} $n$}{
    \textbf{Step 1 (Update the model)}\\
    \For{$j = 1$ \textbf{to} $n'$}{
        Obtain the new data batch ${\bf X}_{j}$ from $D_i$\\
        Update the model on ${\bf X}_{j}$\\
    }
    \textbf{Step 2 (Find the cluster prototype)}\\
    \If{$1 \leq i < n$}{
        $k_i = F_{k}( \rho ) $\\
        ${\bf C}_i = F_{\rm c}(D_i)$ \\
        \ForEach{$c_l \in {\bf C}_i$}{
            $D_i(c_l) = F_{\rm class}(D_i, c_l)$\\
            $D_i(c_l)^* = Divide\text{-}and\text{-}Conquer(D_i(c_l))$\\
            $\{{\bf x}^{c_l}_1, \cdots, {\bf x}^{c_l}_k\} = F_{\rm centre}(D_i(c_l)^*, k)$\\
            \textbf{Step 3 (Accumulate the data samples)}\\
            \For{$h = 1$ \textbf{to} $k$}{
                Obtain ${{\mathcal{M}}(c_l, h)}$ using the \\
                sample selection criterion\\
            }
        }
    }
}
\vspace{1.6pt}
\end{algorithm}
\begin{algorithm}[t]
\caption{Divide-and-Conquer}
\label{alg:DAC}
\textbf{Input:} Sample set $\mathcal{X}$, number of clusters $K$, \\
minimal merged size $m$, recursion depth $d$ \\
\textbf{Output:} Sample subset $\mathcal{X}^*$\\
\If{$k = 0$ \textbf{or} $d = 0$}{\Return $\mathcal{X},\,|\mathcal{X}|$\\}
\textbf{Step 1 (Initial partition)}\\
Compute $K$ clusters with $K$-means: 
$\{\mathcal{C}_1,\dots,\mathcal{C}_K\} \leftarrow {K\text{-means}}(\mathcal{X},\,K)$\\
\textbf{Step 2 (Pairwise distances)}\\
Initialise $D\in\mathbb{R}^{K\times K}$ to $\infty$\\
\For{$1\le a<b\le K$}{
  Compute distance between $\mathcal{C}_a$ \text{and} $\mathcal{C}_b$\\
  $\,D_{a,b}=D_{b,a}\leftarrow\mathrm{Sinkhorn}(\mathcal{C}_a,\mathcal{C}_b)$\\
}
\textbf{Step 3 (Find the closest clusters)}\\
Let $S$ denote all cluster paths with distances in $D$, constrained by the size m\\
$
 s^* \leftarrow \mathrm{find\_shortest\_path}(S,\,m)
$\\
Return $s^*$ with minimal distance and sufficient sample.

\textbf{Step 4 (Merge)}\\
$
  \mathcal{X} \leftarrow \bigcup_{i\in s^*}\mathcal{C}_i
$\\
\textbf{Step 5 (Recurse)}\\
\Return $\mathrm{Divide\text{-}and\text{-}conquer}(\mathcal{X},\,K,\,m,\,d-1)$\\
\end{algorithm}

\section{Additional Experimental Results}
\label{sec:additional_results}

For the experiments, we set the proportion of long-term memory, denoted as $\rho$, to 25\%, 50\%, and 75\%. 
However, since the dataset characteristics and buffer sizes vary, these proportions are approximated using the configurations outlined in \cref{tab:k_approx}.

\begin{table}[h]
  \centering
  \scriptsize
      \caption{Approximation of $k$ across various datasets. ``$k$ (\%)" denotes the number of $k$ values used in ODEDM and the proportion of long-term memory relative to the total buffer size. We can calculate the proportion using $
    |M^{\rm long}_n| + (n-1) \times k \times |c_l| \simeq \lceil |M_n| \times \rho \rceil$, where $n$ is the task number, and $c_l$ denotes the number of classes in a given task.}
  \begin{tabular}{@{}cccc@{}}
  \toprule
  \multicolumn{1}{l}{\textbf{Buffer Size}} & \textbf{CIFAR10} & \textbf{CIFAR100} & \textbf{TINYIMG} \\ \midrule
  \multirow{3}{*}{200}  & 6 (24\%)   & -         & -         \\
                        & 13 (52\%)  & 1 (45\%)  & -         \\
                        & 19 (76\%)  & -         & 1 (90\%)  \\ \midrule
  \multirow{3}{*}{500}  & 16 (26\%)  & 1 (18\%)  & 1 (36\%)  \\
                        & 31 (50\%)  & 3 (54\%)  & -         \\
                        & 47 (75\%)  & 4 (72\%)  & 2 (72\%)  \\ \midrule
  \multirow{3}{*}{5120} & 160 (25\%) & 14 (25\%) & 7 (25\%)  \\
                        & 320 (50\%) & 28 (49\%) & 14 (49\%) \\
                        & 480 (75\%) & 42 (74\%) & 21 (74\%) \\ \bottomrule
  \end{tabular}%
  \label{tab:k_approx}
\end{table}

\begin{table*}[ht]
\centering
\scriptsize
\caption{Performance of DER++refresh on CIFAR10 under varying long-term memory proportions, number of clusters ($K$), and $depth$ in DAC, with a buffer size of 200. Task-IL (Task-Incremental Learning) means the model is informed of the task identity during evaluation.
When evaluating on the $j$-th task, the model only needs to classify among the classes belonging to that task, allowing it to activate the corresponding task-specific head.
Thus, Task-IL primarily measures \textit{retention} of prior tasks rather than inter-task discrimination. Formally, the average accuracy under Task-IL can be written as
$ACC_T^{\text{Task-IL}} = \frac{1}{T} \sum_{j=1}^{T} a_{T,j}^{\text{task}}$,
where $a_{T,j}^{\text{task}}$ denotes the accuracy on the $j$-th task after the model has learned all $T$ tasks, with the task identity provided at test time. }
\begin{tabular}{cccccccc}
\toprule
\multirow{2}{*}{$K$} & \multirow{2}{*}{$depth$} & 
\multicolumn{2}{c}{$25\%$} & 
\multicolumn{2}{c}{$50\%$} & 
\multicolumn{2}{c}{$75\%$} \\
\cmidrule(lr){3-4} \cmidrule(lr){5-6} \cmidrule(lr){7-8}
 & & $ACC_{5}$ & $ACC_5^{\text{Task-IL}}$ & $ACC_{5}$ & $ACC_5^{\text{Task-IL}}$ & $ACC_{5}$ & $ACC_5^{\text{Task-IL}}$ \\
\midrule
3 & 2 & $43.37 \pm 1.81$ & $81.86 \pm 3.06$ & \cellcolor{lightblue!80}{$42.50 \pm 6.19$} & \cellcolor{lightblue!80}{$79.38 \pm 2.42$} & $38.82 \pm 4.38$ & $81.28 \pm 0.83$ \\
3 & 3 & $40.38 \pm 3.16$ & $81.25 \pm 2.26$ & $40.38 \pm 4.90$ & $83.65 \pm 1.35$ & $40.12 \pm 0.75$ & $79.53 \pm 2.40$ \\
3 & 4 & \cellcolor{lightblue!80}{$43.81 \pm 2.25$} & \cellcolor{lightblue!80}{$82.50 \pm 3.72$} & $40.30 \pm 1.00$ & $81.58 \pm 0.79$ & $36.89 \pm 2.46$ & $83.45 \pm 1.21$ \\
4 & 2 & $43.96 \pm 3.12$ & $81.17 \pm 3.64$ & $34.52 \pm 8.06$ & $82.12 \pm 4.40$ & $37.01 \pm 1.89$ & $77.79 \pm 0.72$ \\
4 & 3 & $38.75 \pm 3.52$ & $78.70 \pm 2.76$ & $37.86 \pm 1.01$ & $81.22 \pm 1.84$ & $36.20 \pm 1.75$ & $82.63 \pm 1.54$ \\
4 & 4 & $37.16 \pm 2.67$ & $81.75 \pm 5.73$ & $39.39 \pm 2.63$ & $81.79 \pm 3.12$ & $37.40 \pm 3.32$ & $80.97 \pm 1.30$ \\
5 & 2 & $40.55 \pm 4.39$ & $79.40 \pm 3.89$ & $34.27 \pm 7.25$ & $80.09 \pm 2.41$ & $38.54 \pm 6.16$ & $81.40 \pm 2.82$ \\
5 & 3 & $43.90 \pm 4.94$ & $82.19 \pm 1.70$ & $36.61 \pm 8.25$ & $81.67 \pm 2.23$ & \cellcolor{lightblue!80}{$42.38 \pm 2.03$} & \cellcolor{lightblue!80}{$78.69 \pm 1.93$} \\
5 & 4 & $40.95 \pm 1.91$ & $84.87 \pm 1.23$ & $35.26 \pm 4.57$ & $77.95 \pm 5.66$ & $35.39 \pm 0.65$ & $82.21 \pm 0.83$ \\
\bottomrule
\end{tabular}
\label{tab:proportion}
\end{table*}

\begin{table}[h]
\centering
\scriptsize
\caption{Other OCL baselines on CIFAR10 \cite{wei2023online, buzzega2020dark}, compared with offline training using 50 epochs. $M$ denotes buffer size.}
\label{tab:onpro}
\begin{tabular}{@{}lcc@{}}
\toprule
\multirow{2}{*}{Buffer Method}                                & \multicolumn{2}{c}{CIFAR10 ($ACC_5$)}                     \\ \cmidrule(l){2-3} 
                                                       & $M=200$                & $M=500$                \\ \midrule
\multicolumn{3}{c}{Online training (1 epoch)}                                                                      \\ \midrule
A-GEM \cite{chaudhry2018efficient}                      & $17.5\pm0.3$          & $17.5\pm0.2$          \\
GSS \cite{aljundi2019gradient}                         & $19.4\pm0.7$          & $25.2\pm0.9$          \\
ER \cite{chaudhry2019tiny}                             & $20.9\pm0.9$          & $26.0\pm1.2$          \\
MIR \cite{aljundi2019online}                           & $23.5\pm0.8$          & $29.9\pm1.2$          \\
GDumb \cite{prabhu2020gdumb}                           & $27.1\pm0.7$          & $34.0\pm0.8$          \\
ASER \cite{shim2021online}                             & $22.8\pm0.6$          & $31.6\pm1.1$          \\
CoPE \cite{de2021continual}                            & $37.3\pm2.2$          & $42.9\pm3.5$          \\ 
DER\texttt{++} \cite{buzzega2020dark}                  & $29.7\pm4.2$          & $39.9\pm10.3$          \\ \midrule
\multicolumn{3}{c}{Offline training (50 epochs)}                                                                        \\ \midrule
GEM \cite{lopezpaz2022gradientepisodicmemorycontinual} & $25.54\pm0.76$         & $26.20\pm1.26$         \\
A-GEM \cite{chaudhry2018efficient}                      & $20.04\pm0.34$         & $22.67\pm0.57$         \\
FDR \cite{benjamin2018measuring}                       & $30.91\pm2.74$         & $28.71\pm3.23$         \\ \midrule
\multicolumn{3}{c}{Ours}                                                                        \\ \midrule
DER\texttt{++} w/ ODEDM                                & $\mathbf{42.7}\pm1.9$ & $\mathbf{47.3}\pm2.6$ \\ \bottomrule
\end{tabular}
\end{table}

The best-performing DAC configuration for each memory proportion on CIFAR10 is highlighted in blue in \cref{tab:proportion}. For the 25\% setting, although the configuration with $K=3$ and $depth=4$ achieves an $ACC_{5}$ value comparable to that of $K=4$ and $depth=2$, we identify $K=3$, $depth=4$ as the optimal choice owing to its superior performance in $ACC_{5}^{\text{Task-IL}}$ (see \cref{tab:proportion}). This selected DAC configuration is subsequently adopted for all experiments reported in \cref{tab:accResults}.

\begin{table*}[ht]
\scriptsize
\centering
\caption{$ACC_5$ and $\overline{ACC}_{5}$ of ODEDM on CIFAR10 across three runs under the standard and imbalanced settings with varying buffer sizes. The best results for each setting are indicated in \textbf{bold}.}
\label{tab:cifar10_full}
\begin{tabular}{@{}cccccc@{}}
\toprule
\multicolumn{2}{l}{\multirow{2}{*}{\textbf{Buffer Method}}}  & \multicolumn{2}{c}{\textbf{Standard}}       & \multicolumn{2}{c}{\textbf{Imbalanced}}      \\
\multicolumn{2}{l}{}                                         & $ACC_{5}$        & $\overline{ACC}_{5}$ & $ACC_{5}$         & $\overline{ACC}_{5}$ \\ \midrule
\multirow{20}{*}{$200$}  & DER (ODEDM-6)                     & $\mathbf{29.99}\pm3.48$ & $48.07\pm23.18$  & $14.72\pm3.84$           & $40.97\pm24.01$  \\
                         & DER (ODEDM-13)                    & $26.93\pm3.10$          & $51.06\pm22.16$  & $\mathbf{22.04}\pm5.85$  & $43.82\pm25.43$  \\
                         & DER (ODEDM-19)                    & $22.40\pm1.12$          & $48.25\pm25.17$  & $13.45\pm5.02$           & $40.67\pm21.14$  \\ \cmidrule(l){3-6} 
                         & DER\texttt{++} (ODEDM-6)          & $42.69\pm1.51$          & $54.89\pm13.78$  & $\mathbf{40.95}\pm1.82$  & $57.19\pm18.63$  \\
                         & DER\texttt{++} (ODEDM-13)         & $\mathbf{42.71}\pm1.86$ & $57.35\pm14.51$  & $35.03\pm7.70$           & $51.87\pm14.10$  \\
                         & DER\texttt{++} (ODEDM-19)         & $38.69\pm0.69$          & $58.52\pm16.48$  & $31.81\pm1.83$           & $53.54\pm18.86$  \\ \cmidrule(l){3-6} 
                         & DER\texttt{++}refresh (ODEDM-6)   & $41.63\pm1.75$          & $54.14\pm14.27$  & $37.56\pm3.08$           & $54.45\pm17.35$  \\
                         & DER\texttt{++}refresh (ODEDM-13)  & $\mathbf{42.47}\pm0.38$ & $57.91\pm13.60$  & $\mathbf{38.41}\pm0.88$  & $52.12\pm14.12$  \\
                         & DER\texttt{++}refresh (ODEDM-19)  & $36.19\pm4.73$          & $56.21\pm17.95$  & $35.88\pm1.67$           & $53.90\pm16.42$  \\ \cmidrule(l){3-6} 
                         & FDR (ODEDM-6)                     & $21.15\pm5.54$          & $38.76\pm17.57$  & $\mathbf{21.14}\pm1.72$  & $40.15\pm21.32$  \\
                         & FDR (ODEDM-13)                    & $17.56\pm2.85$          & $42.19\pm24.82$  & $20.18\pm2.53$           & $36.79\pm21.18$  \\
                         & FDR (ODEDM-19)                    & $\mathbf{25.17}\pm1.39$ & $43.30\pm25.82$  & $16.88\pm2.15$           & $38.74\pm23.03$  \\ \cmidrule(l){3-6} 
                         & iCaRL (ODEDM-6)                   & $\mathbf{37.64}\pm1.56$ & $58.58\pm19.40$  & $\mathbf{35.26}\pm0.58$  & $56.22\pm18.40$  \\
                         & iCaRL (ODEDM-13)                  & $36.65\pm2.96$          & $56.80\pm18.60$  & $32.21\pm2.71$           & $52.49\pm19.13$  \\
                         & iCaRL (ODEDM-19)                  & $31.42\pm2.32$          & $51.42\pm18.46$  & $29.93\pm1.34$           & $51.46\pm20.69$  \\ \cmidrule(l){3-6} 
                         & PuriDivER (ODEDM-6)               & $20.77\pm1.04$          & $42.78\pm24.60$  & $19.89\pm0.34$           & $41.33\pm23.13$  \\
                         & PuriDivER (ODEDM-13)              & $\mathbf{23.04}\pm1.30$ & $42.90\pm23.02$  & $\mathbf{22.08}\pm1.13$  & $41.08\pm23.89$  \\
                         & PuriDivER (ODEDM-19)              & $22.20\pm0.37$          & $43.03\pm23.99$  & $21.56\pm1.81$           & $43.20\pm23.76$  \\ \midrule
\multirow{22}{*}{$500$}  & DER (ODEDM-16)                    & $24.98\pm5.26$          & $47.65\pm22.04$  & $20.37\pm5.56$           & $41.88\pm22.50$  \\
                         & DER (ODEDM-31)                    & $\mathbf{28.34}\pm4.98$ & $48.41\pm20.53$  & $19.60\pm2.65$           & $41.60\pm23.59$  \\
                         & DER (ODEDM-47)                    & $23.18\pm3.64$          & $44.94\pm22.47$  & $\mathbf{21.32}\pm5.02$  & $40.49\pm21.07$  \\ \cmidrule(l){3-6} 
                         & DER\texttt{++} (ODEDM-16)         & $\mathbf{47.31}\pm2.62$ & $56.02\pm9.82$  & $\mathbf{40.14}\pm2.46$  & $59.03\pm17.23$  \\
                         & DER\texttt{++} (ODEDM-31)         & $45.53\pm3.39$          & $56.57\pm8.41$  & $39.51\pm4.05$           & $50.57\pm16.73$  \\
                         & DER\texttt{++} (ODEDM-47)         & $44.67\pm1.08$          & $55.36\pm16.73$  & $36.69\pm3.99$           & $52.79\pm19.17$  \\ \cmidrule(l){3-6} 
                         & DER\texttt{++}refresh (ODEDM-16)  & $\mathbf{46.17}\pm2.37$ & $60.04\pm14.36$  & $29.11\pm10.39$          & $51.60\pm15.99$  \\
                         & DER\texttt{++}refresh (ODEDM-31)  & $38.30\pm11.61$         & $59.00\pm19.97$  & $\mathbf{41.53}\pm4.15$  & $54.06\pm15.56$  \\
                         & DER\texttt{++}refresh (ODEDM-47)  & $44.21\pm0.61$          & $55.20\pm11.71$  & $40.38\pm1.94$           & $52.74\pm18.83$  \\ \cmidrule(l){3-6} 
                         & FDR (ODEDM-16)                    & $\mathbf{24.43}\pm1.04$ & $44.45\pm25.30$  & $16.59\pm1.16$           & $34.01\pm17.48$  \\
                         & FDR (ODEDM-31)                    & $19.52\pm2.91$          & $40.89\pm25.73$  & $16.75\pm1.59$           & $37.17\pm21.36$  \\
                         & FDR (ODEDM-47)                    & $15.78\pm1.37$          & $39.26\pm23.29$  & $\mathbf{17.50}\pm1.57$  & $32.94\pm18.89$  \\ \cmidrule(l){3-6} 
                         & iCaRL (ODEDM-16)                  & $35.42\pm4.39$          & $54.00\pm18.02$  & $35.56\pm2.25$           & $55.59\pm16.99$  \\
                         & iCaRL (ODEDM-31)                  & $\mathbf{38.31}\pm0.72$ & $55.74\pm17.55$  & $34.70\pm1.56$           & $53.58\pm17.09$  \\
                         & iCaRL (ODEDM-47)                  & $35.04\pm2.75$          & $56.14\pm19.11$  & $\mathbf{36.10}\pm0.87$  & $55.64\pm18.66$  \\ \cmidrule(l){3-6} 
                         & PuriDivER (ODEDM-16)              & $23.21\pm0.81$          & $43.81\pm23.79$  & $20.84\pm0.65$           & $42.84\pm22.92$  \\
                         & PuriDivER (ODEDM-31)              & $21.49\pm1.08$          & $44.25\pm24.59$  & $20.84\pm2.77$           & $41.16\pm23.05$  \\
                         & PuriDivER (ODEDM-47)              & $\mathbf{23.73}\pm1.83$ & $42.97\pm23.19$  & $\mathbf{22.76}\pm1.26$  & $43.03\pm23.78$  \\ \midrule
\multirow{18}{*}{$5120$} & DER (ODEDM-160)                   & $21.81\pm4.02$          & $48.67\pm24.35$  & $18.94\pm5.34$           & $38.91\pm21.56$  \\
                         & DER (ODEDM-320)                   & $\mathbf{29.96}\pm1.63$ & $46.82\pm19.38$  & $\mathbf{21.89}\pm4.60$  & $40.51\pm25.19$  \\
                         & DER (ODEDM-480)                   & $26.34\pm3.87$          & $46.12\pm19.43$  & $17.91\pm0.73$           & $38.37\pm20.94$  \\ \cmidrule(l){3-6} 
                         & DER\texttt{++} (ODEDM-160)        & $\mathbf{54.49}\pm1.44$ & $60.36\pm10.89$  & $43.37\pm3.58$           & $52.47\pm8.85$  \\
                         & DER\texttt{++} (ODEDM-320)        & $49.81\pm9.49$          & $64.15\pm17.56$  & $44.34\pm5.27$           & $54.57\pm15.04$  \\
                         & DER\texttt{++} (ODEDM-480)        & $48.75\pm10.79$         & $60.00\pm12.54$  & $\mathbf{47.90}\pm2.10$  & $53.82\pm11.45$  \\ \cmidrule(l){3-6} 
                         & DER\texttt{++}refresh (ODEDM-160) & $55.14\pm2.30$          & $63.14\pm15.17$  & $\mathbf{45.22}\pm4.63$  & $51.49\pm14.41$  \\
                         & DER\texttt{++}refresh (ODEDM-320) & $\mathbf{55.17}\pm2.30$ & $62.97\pm12.66$  & $43.68\pm1.83$           & $50.89\pm12.21$  \\
                         & DER\texttt{++}refresh (ODEDM-480) & $52.87\pm1.06$          & $63.77\pm12.51$  & $42.39\pm0.88$           & $52.70\pm14.37$  \\ \cmidrule(l){3-6} 
                         & FDR (ODEDM-160)                   & $16.84\pm0.75$          & $39.74\pm26.70$  & $17.26\pm1.35$           & $39.27\pm27.09$  \\
                         & FDR (ODEDM-320)                   & $\mathbf{18.33}\pm0.58$ & $38.99\pm24.61$  & $\mathbf{17.86}\pm0.14$  & $35.77\pm22.49$  \\
                         & FDR (ODEDM-480)                   & $17.48\pm0.78$          & $39.22\pm24.19$  & $16.00\pm0.00$           & $40.79\pm28.22$  \\ \cmidrule(l){3-6} 
                         & iCaRL (ODEDM-160)                 & $\mathbf{47.50}\pm0.88$ & $64.95\pm14.69$  & $\mathbf{44.78}\pm2.11$  & $60.86\pm15.84$  \\
                         & iCaRL (ODEDM-320)                 & $44.43\pm1.20$          & $62.16\pm17.43$  & $38.94\pm2.82$           & $58.36\pm18.25$  \\
                         & iCaRL (ODEDM-480)                 & $43.72\pm1.43$          & $63.08\pm17.13$  & $41.79\pm1.91$           & $58.49\pm16.85$  \\ \bottomrule
\end{tabular}%
\end{table*}

\begin{table*}[ht]
\scriptsize
\centering
\begin{tabular}{@{}cccccc@{}}
\toprule
\multicolumn{2}{l}{\multirow{2}{*}{\textbf{Buffer Method}}}  & \multicolumn{2}{c}{\textbf{Standard}}       & \multicolumn{2}{c}{\textbf{Imbalanced}}      \\
\multicolumn{2}{l}{}                                         & $ACC_{5}$        & $\overline{ACC}_{5}$ & $ACC_{5}$         & $\overline{ACC}_{5}$ \\ \midrule
\multirow{6}{*}{$200$}  
                         & VR-MCL (ODEDM-6)                   & $40.61\pm0.08$ & $60.50\pm17.00$  & 
                         $41.16\pm1.66$ & $55.99\pm16.92$  \\
                         & VR-MCL (ODEDM-13)                  & 
                         $\mathbf{41.34}\pm4.79$          & $56.31\pm17.71$  & 
                         $\mathbf{42.51}\pm1.37$           & $57.88\pm17.49$  \\
                         & VR-MCL (ODEDM-19)                  & 
                         $40.20\pm4.99$          & $60.70\pm18.12$  & 
                         $34.26\pm2.36$           & $54.72\pm19.35$  \\ \cmidrule(l){3-6} 
                         & POCL (ODEDM-6)               & 
                         $\mathbf{37.03}\pm5.19$          & $53.85\pm15.28$  & 
                         $\mathbf{36.90}\pm4.76$           & $56.29\pm17.37$  \\
                         & POCL (ODEDM-13)              & 
                         $32.48\pm1.73$ & $54.82\pm20.66$  & 
                         $24.74\pm4.44$  & $52.24\pm23.81$  \\
                         & POCL (ODEDM-19)              & 
                         $28.67\pm5.22$          & $54.91\pm22.92$  & 
                         $26.96\pm3.25$           & $49.81\pm17.80$  \\ \midrule
\multirow{6}{*}{$500$}
                         & VR-MCL (ODEDM-16)                  & 
                         $\mathbf{47.50}\pm1.93$ & $63.13\pm17.15$ & 
                         $46.14\pm2.73$ & $60.56\pm14.15$  \\
                         & VR-MCL (ODEDM-31)                  & 
                         $46.78\pm2.04$ & $58.36\pm14.50$ & 
                         $\mathbf{47.70}\pm3.55$ & $55.32\pm9.22$  \\
                         & VR-MCL (ODEDM-47)                  & 
                         $46.30\pm2.77$ & $59.50\pm12.62$  & 
                         $39.99\pm0.49$ & $56.06\pm16.49$  \\ \cmidrule(l){3-6} 
                         & POCL (ODEDM-16) & 
                         $\mathbf{48.48}\pm0.94$ & $64.05\pm13.00$  & 
                         $\mathbf{44.29}\pm2.09$ & $60.80\pm15.53$  \\
                         & POCL (ODEDM-31) & 
                         $44.10\pm2.76$ & $62.79\pm16.14$  & 
                         $43.95\pm3.54$ & $59.82\pm14.08$  \\
                         & POCL (ODEDM-47) & 
                         $32.36\pm8.71$ & $60.77\pm19.74$  & 
                         $33.53\pm2.69$ & $53.73\pm15.75$  \\ \midrule
\multirow{6}{*}{$5120$}
                         & VR-MCL (ODEDM-160)                   & 
                         $\mathbf{55.39}\pm0.56$ & $68.03\pm13.57$  & 
                         $47.54\pm0.97$ & $57.50\pm14.10$  \\
                         & VR-MCL (ODEDM-320)                   & 
                         $51.15\pm2.67$ & $59.90\pm15.68$  & 
                         $\mathbf{47.70}\pm3.23$ & $56.52\pm14.14$  \\
                         & VR-MCL (ODEDM-480)                   & 
                         $51.47\pm1.29$ & $60.41\pm10.76$  & 
                         $47.61\pm2.52$ & $53.63\pm8.50$  \\ \cmidrule(l){3-6} 
                         & POCL (ODEDM-160)                 & 
                         $\mathbf{66.74}\pm1.13$ & $74.38\pm7.52$ & 
                         $\mathbf{62.01}\pm3.15$ & $63.99\pm5.81$  \\
                         & POCL (ODEDM-320)                 & 
                         $64.66\pm3.73$ & $76.44\pm10.07$  & 
                         $61.18\pm4.30$ & $68.16\pm9.07$  \\
                         & POCL (ODEDM-480)                 & 
                         $60.73\pm4.17$ & $70.00\pm9.15$  & 
                         $60.06\pm2.00$ & $67.71\pm8.89$  \\ \bottomrule
\end{tabular}%
\end{table*}

\begin{table*}[ht]
\scriptsize
\centering
\caption{$ACC_{10}$ and $\overline{ACC}_{10}$ of ODEDM on CIFAR100 across three runs under the standard and imbalanced settings with varying buffer sizes. The best results for each setting are indicated in \textbf{bold}.}
\label{tab:cifar100_full}
\begin{tabular}{@{}cccccc@{}}
\toprule
\multicolumn{2}{l}{\multirow{2}{*}{\textbf{Buffer Method}}} & \multicolumn{2}{c}{\textbf{Standard}}       & \multicolumn{2}{c}{\textbf{Imbalanced}}     \\
\multicolumn{2}{l}{}                                        & $ACC_{10}$        & $\overline{ACC}_{10}$ & $ACC_{10}$        & $\overline{ACC}_{10}$ \\ \midrule
\multirow{8}{*}{$200$}   & DER (ODEDM-1)                    & $4.78\pm0.56$          & $10.18\pm7.50$  & $3.20\pm0.71$          & $8.29\pm7.39$  \\
                         & DER\texttt{++} (ODEDM-1)         & $7.78\pm1.78$          & $16.15\pm8.92$  & $7.46\pm1.57$          & $12.51\pm6.68$  \\
                         & DER\texttt{++}refresh (ODEDM-1)  & $8.47\pm1.24$          & $15.30\pm8.44$  & $5.68\pm1.99$          & $12.51\pm7.57$  \\
                         & FDR (ODEDM-1)                    & $4.14\pm0.52$          & $12.43\pm9.49$  & $3.49\pm0.55$          & $8.92\pm6.42$  \\
                         & iCaRL (ODEDM-1)                  & $1.03\pm0.02$          & $13.53\pm14.02$ & $1.17\pm0.34$          & $12.23\pm12.13$  \\
                         & PuriDivER (ODEDM-1)              & $7.67\pm0.23$          & $14.67\pm7.94$  & $6.83\pm0.43$          & $13.32\pm6.56$  \\
                         & VR-MCL (ODEDM-1)                 & 
                         $15.31\pm0.70$ & $25.47\pm10.77$  & 
                         $12.84\pm0.40$ & $21.52\pm10.00$  \\
                         & POCL (ODEDM-1)                   & 
                         $9.16\pm0.34$ & $20.43\pm9.03$  & 
                         $8.31\pm0.68$ & $17.74\pm7.50$  \\ \midrule
\multirow{28}{*}{$500$}  & DER (ODEDM-1)                    & $\mathbf{4.12}\pm0.81$ & $10.52\pm7.93$  & $3.61\pm1.09$          & $7.30\pm4.06$  \\
                         & DER (ODEDM-3)                    & $3.63\pm0.38$          & $10.49\pm7.24$  & $3.75\pm0.54$          & $9.05\pm7.53$  \\
                         & DER (ODEDM-4)                    & $4.10\pm0.89$          & $10.11\pm7.13$  & $\mathbf{3.84}\pm0.34$ & $8.29\pm6.13$  \\ \cmidrule(l){3-6} 
                         & DER\texttt{++} (ODEDM-1)         & $10.77\pm3.76$          & $17.57\pm6.97$  & $\mathbf{10.16}\pm0.55$ & $14.27\pm6.46$  \\
                         & DER\texttt{++} (ODEDM-3)         & $\mathbf{13.62}\pm2.51$ & $18.37\pm7.61$  & $8.16\pm0.99$          & $15.28\pm7.08$  \\
                         & DER\texttt{++} (ODEDM-4)         & $11.30\pm0.59$          & $19.03\pm7.47$  & $7.53\pm0.87$          & $15.57\pm9.29$  \\ \cmidrule(l){3-6} 
                         & DER\texttt{++}refresh (ODEDM-1)  & $9.11\pm3.63$          & $20.43\pm11.24$ & $\mathbf{10.27}\pm1.19$ & $14.16\pm6.77$  \\
                         & DER\texttt{++}refresh (ODEDM-3)  & $8.93\pm1.19$          & $19.26\pm7.46$  & $9.69\pm2.35$          & $16.04\pm5.93$  \\
                         & DER\texttt{++}refresh (ODEDM-4)  & $\mathbf{11.79}\pm1.91$ & $17.78\pm5.93$  & $8.26\pm3.15$          & $15.67\pm8.10$  \\ \cmidrule(l){3-6} 
                         & FDR (ODEDM-1)                    & $4.73\pm0.81$          & $11.74\pm8.61$  & $3.96\pm0.45$          & $10.09\pm8.47$  \\
                         & FDR (ODEDM-3)                    & $4.90\pm0.40$          & $12.49\pm9.78$  & $4.09\pm0.69$          & $10.01\pm7.56$  \\
                         & FDR (ODEDM-4)                    & $\mathbf{4.96}\pm0.84$ & $12.20\pm7.90$  & $\mathbf{4.71}\pm0.66$ & $10.22\pm7.42$  \\ \cmidrule(l){3-6} 
                         & iCaRL (ODEDM-1)                  & $\mathbf{7.35}\pm0.14$ & $16.62\pm9.84$  & $\mathbf{7.49}\pm0.40$ & $15.44\pm8.40$  \\
                         & iCaRL (ODEDM-3)                  & $0.95\pm0.08$          & $14.52\pm11.93$ & $0.99\pm0.19$          & $12.92\pm10.96$  \\
                         & iCaRL (ODEDM-4)                  & $1.07\pm0.18$          & $13.75\pm10.93$ & $1.17\pm0.12$          & $14.15\pm12.44$  \\ \cmidrule(l){3-6} 
                         & PuriDivER (ODEDM-1)              & $7.65\pm0.22$          & $15.20\pm8.06$  & $\mathbf{7.61}\pm0.39$ & $12.85\pm5.68$  \\
                         & PuriDivER (ODEDM-3)              & $\mathbf{8.04}\pm0.39$ & $15.57\pm7.89$  & $6.90\pm0.65$          & $13.22\pm6.41$  \\
                         & PuriDivER (ODEDM-4)              & $7.84\pm0.36$          & $15.99\pm8.65$  & $6.38\pm0.25$          & $12.40\pm5.82$  \\ \cmidrule(l){3-6}
                         & VR-MCL (ODEDM-1)                 & 
                         $\mathbf{18.32}\pm1.07$ & $27.35\pm9.20$  & 
                         $15.00\pm0.81$ & $22.42\pm8.02$  \\
                         & VR-MCL (ODEDM-3)                 & 
                         $18.31\pm0.67$ & $28.54\pm9.83$  & 
                         $\mathbf{16.54}\pm0.47$ & $24.09\pm8.28$  \\
                         & VR-MCL (ODEDM-4)                 & 
                         $18.30\pm0.55$ & $25.86\pm7.33$  & 
                         $15.90\pm0.49$ & $22.51\pm6.78$  \\ \cmidrule(l){3-6} 
                         & POCL (ODEDM-1)                   & 
                         $\mathbf{15.82}\pm1.48$ & $25.39\pm7.32$  & 
                         $13.55\pm0.72$ & $20.59\pm5.73$   \\
                         & POCL (ODEDM-3)                   & 
                         $13.95\pm0.83$ & $23.24\pm7.19$  & 
                         $\mathbf{14.03}\pm0.65$ & $23.32\pm6.67$  \\
                         & POCL (ODEDM-4)                   & 
                         $11.87\pm0.91$ & $21.98\pm8.31$   & 
                         $11.27\pm0.73$ & $20.26\pm6.44$  \\ \midrule
\multirow{25}{*}{$5120$} & DER (ODEDM-14)                   & $3.46\pm0.15$          & $11.59\pm9.57$  & $\mathbf{2.92}\pm0.78$ & $8.03\pm5.39$  \\
                         & DER (ODEDM-28)                   & $\mathbf{4.83}\pm0.33$ & $11.23\pm10.10$ & $2.73\pm0.48$          & $7.92\pm5.72$  \\
                         & DER (ODEDM-42)                   & $3.92\pm0.74$          & $10.47\pm7.56$  & $2.54\pm0.97$          & $8.30\pm6.44$  \\ \cmidrule(l){3-6} 
                         & DER\texttt{++} (ODEDM-14)        & $\mathbf{22.42}\pm5.67$ & $24.84\pm6.90$  & $11.53\pm2.53$          & $16.16\pm5.61$  \\
                         & DER\texttt{++} (ODEDM-28)        & $16.47\pm1.28$          & $24.63\pm7.14$  & $\mathbf{13.62}\pm3.94$ & $17.55\pm6.00$  \\
                         & DER\texttt{++} (ODEDM-42)        & $15.04\pm3.61$          & $23.95\pm7.15$  & $12.38\pm5.90$          & $16.41\pm6.94$  \\ \cmidrule(l){3-6} 
                         & DER\texttt{++}refresh (ODEDM-14) & $15.11\pm2.96$          & $24.34\pm8.22$  & $\mathbf{15.15}\pm1.92$ & $18.60\pm8.99$  \\
                         & DER\texttt{++}refresh (ODEDM-28) & $\mathbf{21.60}\pm3.27$ & $24.46\pm6.93$  & $11.84\pm4.73$          & $15.95\pm7.02$  \\
                         & DER\texttt{++}refresh (ODEDM-42) & $18.73\pm6.21$          & $24.28\pm9.02$  & $13.30\pm4.21$          & $18.28\pm5.58$  \\ \cmidrule(l){3-6} 
                         & FDR (ODEDM-14)                   & $\mathbf{6.04}\pm0.94$ & $13.84\pm10.31$ & $4.76\pm0.64$          & $9.96\pm6.61$  \\
                         & FDR (ODEDM-28)                   & $5.17\pm0.72$          & $12.39\pm7.88$  & $4.60\pm0.26$          & $10.83\pm9.41$  \\
                         & FDR (ODEDM-42)                   & $4.98\pm0.76$          & $12.21\pm8.84$  & $\mathbf{4.94}\pm0.12$ & $11.14\pm7.98$  \\ \cmidrule(l){3-6} 
                         & iCaRL (ODEDM-14)                 & $8.49\pm0.35$          & $16.67\pm7.74$  & $7.58\pm0.54$          & $14.93\pm7.72$  \\
                         & iCaRL (ODEDM-28)                 & $\mathbf{9.28}\pm0.14$ & $15.95\pm7.15$  & $\mathbf{8.33}\pm0.16$ & $15.51\pm7.90$  \\
                         & iCaRL (ODEDM-42)                 & $8.45\pm0.52$          & $17.21\pm8.92$  & $7.94\pm0.23$          & $15.97\pm8.32$ \\ \cmidrule(l){3-6}
                         & VR-MCL (ODEDM-14)                & 
                         $24.72\pm1.03$ & $28.81\pm5.80$  & 
                         $22.20\pm1.21$ & $27.79\pm8.53$   \\ 
                         & VR-MCL (ODEDM-28)                & 
                         $\mathbf{25.37}\pm0.97$ & $31.19\pm6.60$  & 
                         $21.08\pm0.56$ & $26.06\pm6.43$  \\
                         & VR-MCL (ODEDM-42)                & 
                         $24.94\pm0.28$ & $30.52\pm6.91$  & 
                         $\mathbf{22.41}\pm0.99$ & $27.48\pm6.17$  \\ \cmidrule(l){3-6} 
                         & POCL (ODEDM-14)                  & 
                         $\mathbf{35.56}\pm0.61$ & $36.34\pm4.23$  & 
                         $\mathbf{29.56}\pm4.06$ & $27.75\pm4.91$  \\
                         & POCL (ODEDM-28)                  & 
                         $33.13\pm1.49$ & $34.52\pm3.97$  & 
                         $26.65\pm3.05$ & $25.93\pm4.10$  \\
                         & POCL (ODEDM-42)                  & 
                         $34.29\pm1.76$ & $36.26\pm5.41$  & 
                         $29.24\pm2.04$ & $27.46\pm4.45$ \\ \bottomrule
\end{tabular}
\end{table*}

\begin{table*}[ht]
\scriptsize
\centering
\caption{$ACC_{10}$ and $\overline{ACC}_{10}$ of ODEDM on TINYIMG across three runs under the standard and imbalanced settings with varying buffer sizes. The best results for each setting are indicated in \textbf{bold}.}
\label{tab:tinyimg_full}
\begin{tabular}{@{}cccccc@{}}
\toprule
\multicolumn{2}{l}{\multirow{2}{*}{\textbf{Buffer Method}}} & \multicolumn{2}{c}{\textbf{Standard}}       & \multicolumn{2}{c}{\textbf{Imbalanced}}     \\ [0.2em] 
\multicolumn{2}{l}{}                                        & $ACC_{10}$        & $\overline{ACC}_{10}$ & $ACC_{10}$        & $\overline{ACC}_{10}$ \\ \midrule
\multirow{7}{*}{$200$}   & DER (ODEDM-1)                    & $3.83\pm0.27$          & $8.90\pm6.02$  & $2.82\pm0.45$          & $6.68\pm4.86$  \\
                         & DER\texttt{++} (ODEDM-1)         & $4.65\pm1.16$          & $12.14\pm8.30$  & $4.09\pm0.95$          & $8.90\pm5.50$  \\
                         & DER\texttt{++}refresh (ODEDM-1)  & $5.25\pm1.09$          & $12.85\pm8.20$  & $3.29\pm0.73$          & $9.45\pm6.32$  \\
                         & FDR (ODEDM-1)                    & $4.18\pm0.31$          & $10.94\pm8.13$  & $3.51\pm0.50$          & $8.56\pm5.85$  \\
                         & iCaRL (ODEDM-1)                  & $-$                      & $-$              & $-$                      & $-$              \\
                         & VR-MCL (ODEDM-1)                  & 
                         $8.64\pm0.33$ & $16.67\pm8.50$ & 
                         $7.70\pm0.04$ & $13.28\pm5.94$ \\
                         & POCL (ODEDM-1)                  & 
                         $4.77\pm0.56$ & $11.33\pm6.63$ & 
                         $3.41\pm0.52$ & $9.56\pm5.00$  \\ \midrule
\multirow{18}{*}{$500$}  & DER (ODEDM-1)                    & $\mathbf{4.45}\pm0.12$ & $9.38\pm6.72$  & $\mathbf{3.55}\pm0.17$ & $6.60\pm4.31$  \\
                         & DER (ODEDM-2)                    & $3.58\pm0.40$          & $9.62\pm7.82$  & $3.26\pm0.57$          & $7.18\pm5.09$  \\ \cmidrule(l){3-6} 
                         & DER\texttt{++} (ODEDM-1)         & $\mathbf{7.43}\pm0.75$ & $14.34\pm6.10$  & $5.56\pm0.64$          & $10.08\pm5.12$  \\
                         & DER\texttt{++} (ODEDM-2)         & $6.90\pm0.65$          & $14.29\pm6.91$  & $\mathbf{5.93}\pm1.42$ & $12.49\pm6.24$  \\ \cmidrule(l){3-6} 
                         & DER\texttt{++}refresh (ODEDM-1)  & $\mathbf{7.14}\pm0.85$ & $13.10\pm6.37$  & $\mathbf{7.00}\pm0.79$ & $10.77\pm6.06$  \\
                         & DER\texttt{++}refresh (ODEDM-2)  & $5.76\pm0.47$          & $13.50\pm6.89$  & $5.63\pm0.66$          & $11.27\pm4.75$  \\ \cmidrule(l){3-6} 
                         & FDR (ODEDM-1)                    & $\mathbf{4.84}\pm0.38$ & $10.76\pm7.45$  & $3.47\pm0.11$          & $8.56\pm5.73$  \\
                         & FDR (ODEDM-2)                    & $4.66\pm0.42$          & $11.26\pm8.76$  & $\mathbf{3.90}\pm0.27$ & $8.48\pm5.77$  \\ \cmidrule(l){3-6} 
                         & iCaRL (ODEDM-1)                  & $0.44\pm0.09$          & $6.69\pm6.78$  & $0.45\pm0.03$          & $6.84\pm6.89$  \\
                         & iCaRL (ODEDM-2)                  & $\mathbf{0.48}\pm0.08$ & $7.17\pm8.14$  & $\mathbf{0.51}\pm0.12$ & $5.91\pm6.81$  \\ \cmidrule(l){3-6} 
                         & VR-MCL (ODEDM-1)                    & 
                         $10.61\pm0.52$ & $17.45\pm7.52$  & 
                         $\mathbf{10.60}\pm0.82$ & $16.13\pm5.41$ \\
                         & VR-MCL (ODEDM-2)                    & 
                         $\mathbf{10.74}\pm0.59$ & $18.22\pm7.08$  & 
                         $9.34\pm0.53$ & $16.74\pm6.73$  \\ \cmidrule(l){3-6} 
                         & POCL (ODEDM-1)                  & 
                         $\mathbf{6.91}\pm0.46$ & $14.07\pm6.38$  & 
                         $\mathbf{6.47}\pm0.27$ & $11.29\pm4.20$  \\
                         & POCL (ODEDM-2)                  & 
                         $6.76\pm0.78$ & $14.67\pm6.54$  & 
                         $5.70\pm0.28$ & $11.65\pm4.80$  \\ \midrule
\multirow{25}{*}{$5120$} 
                         & DER (ODEDM-7)                   & $4.37\pm0.49$          & $8.84\pm5.24$  & $3.27\pm0.36$ & $7.02\pm5.02$  \\
                         & DER (ODEDM-14)                   & $4.32\pm0.12$          & $9.58\pm6.67$  & $\mathbf{3.55}\pm0.37$ & $7.28\pm4.40$  \\
                         & DER (ODEDM-21)                   & $\mathbf{4.49}\pm0.48$ & $9.29\pm5.78$  & $3.22\pm0.27$          & $7.57\pm5.74$  \\ \cmidrule(l){3-6} 
                         & DER\texttt{++} (ODEDM-7)        & $\mathbf{15.61}\pm3.29$          & $20.09\pm4.61$  & $9.67\pm2.86$ & $15.51\pm5.21$  \\
                         & DER\texttt{++} (ODEDM-14)        & $13.38\pm1.09$          & $20.38\pm5.09$  & $\mathbf{11.13}\pm1.72$ & $15.11\pm3.65$  \\
                         & DER\texttt{++} (ODEDM-21)        & $14.21\pm1.16$ & $20.69\pm5.96$  & $10.35\pm1.54$          & $16.29\pm3.97$  \\ \cmidrule(l){3-6} 
                         & DER\texttt{++}refresh (ODEDM-7) & $15.88\pm1.37$          & $19.90\pm3.82$  & $\mathbf{12.71}\pm1.17$ & $15.49\pm3.61$  \\
                         & DER\texttt{++}refresh (ODEDM-14) & $17.27\pm2.20$          & $19.71\pm4.31$  & $11.92\pm2.79$ & $15.70\pm3.73$  \\
                         & DER\texttt{++}refresh (ODEDM-21) & $\mathbf{17.65}\pm0.43$ & $21.97\pm5.78$  & $10.83\pm1.01$          & $16.16\pm3.79$  \\ \cmidrule(l){3-6} 
                         & FDR (ODEDM-7)                   & $4.79\pm0.39$          & $11.07\pm8.39$  & $4.22\pm0.33$          & $9.09\pm5.82$  \\
                         & FDR (ODEDM-14)                   & $4.77\pm0.66$          & $11.44\pm7.43$  & $4.53\pm0.12$          & $9.03\pm5.50$  \\
                         & FDR (ODEDM-21)                   & $\mathbf{5.01}\pm0.23$ & $11.88\pm7.48$  & $\mathbf{4.69}\pm0.22$ & $8.84\pm5.40$  \\ \cmidrule(l){3-6} 
                         & iCaRL (ODEDM-7)                 & $3.97\pm0.14$ & $8.13\pm4.84$  & $\mathbf{3.85}\pm0.18$ & $8.63\pm5.83$  \\
                         & iCaRL (ODEDM-14)                 & $\mathbf{4.01}\pm0.02$ & $9.61\pm6.72$  & $3.68\pm0.08$ & $8.16\pm5.33$  \\
                         & iCaRL (ODEDM-21)                 & $3.81\pm0.11$          & $8.76\pm5.42$  & $3.37\pm0.20$          & $8.31\pm5.46$  \\ \cmidrule(l){3-6} 
                         & VR-MCL (ODEDM-7)                   & 
                         $18.20\pm0.90$ & $21.58\pm5.34$  & 
                         $16.02\pm0.54$ & $20.33\pm5.43$  \\
                         & VR-MCL (ODEDM-14)                   & 
                         $18.65\pm0.38$ & $21.77\pm3.45$  & 
                         $\mathbf{17.49}\pm0.78$ & $19.71\pm3.49$  \\
                         & VR-MCL (ODEDM-21)                   & 
                         $\mathbf{19.69}\pm0.31$ & $24.18\pm5.36$  & 
                         $16.36\pm0.94$ & $20.21\pm4.75$  \\ \cmidrule(l){3-6} 
                         & POCL (ODEDM-7)                 & 
                         $\mathbf{21.15}\pm0.62$ & $24.26\pm3.07$  & 
                         $19.23\pm0.62$ & $21.60\pm3.64$  \\
                         & POCL (ODEDM-14)                 & 
                         $19.37\pm0.53$ & $24.68\pm3.67$  & 
                         $19.36\pm1.05$ & $21.99\pm4.20$  \\
                         & POCL (ODEDM-21)                 & 
                         $19.62\pm0.34$ & $25.07\pm3.47$  & 
                         $\mathbf{19.58}\pm0.24$ & $22.21\pm2.61$  \\ \bottomrule
\end{tabular}%
\end{table*}

\subsection{PuriDivER with Buffer Fitting}
PuriDivER can operate in two modes: with buffer fitting and without buffer fitting. Buffer fitting involves training the model for several additional rounds on the samples stored in the memory buffer at the end of each task. For a fair comparison, the results reported in \cref{tab:accResults} exclude buffer fitting. In this section, we instead examine the performance of PuriDivER with and without ODEDM when buffer fitting is enabled.

Buffer fitting is structured into three sequential stages~\cite{bang2022online}. It begins with a warm-up phase, conducted over a predefined number of epochs, in which the model is trained solely on the buffer samples using the cross-entropy loss. This initial phase is intended to stabilise the optimisation dynamics before more advanced procedures are introduced. \\

Following the warm-up, the procedure transitions into the PuriDivER stage, where buffer samples are categorised into three groups: (i) clean samples, (ii) ambiguous samples, and (iii) incorrect samples. When this partitioning is reliable, training is performed using the MixMatch framework~\cite{berthelot2019mixmatchholisticapproachsemisupervised}, in which clean samples are treated as labelled data, while ambiguous and incorrect samples are considered as unlabeled or soft-labelled data. Finally, if the partitioning is deemed unreliable, due to an insufficient number of ambiguous or incorrect samples, the algorithm defaults to a fallback strategy. In this case, training continues on buffer samples using only the cross-entropy loss.

\cref{tab:PuriDivER} demonstrates that enlarging the buffer size from 200 to 500 yields substantial accuracy improvements on the standard setting. In particular, with a buffer size of 500, PuriDivER with ODEDM attains 61.45\% in $ACC_5$ and 70.53\% in $\overline{ACC}_{5}$, all of which outperform the corresponding PuriDivER baseline. By contrast, for the imbalanced setting, PuriDivER with ODEDM degrades performance, especially when the buffer size is 500. The results indicate that augmenting PuriDivER with ODEDM yields performance gains during buffer fitting, although these improvements are observed only under the standard setting.

\begin{table*}[ht]
\centering
\scriptsize
\caption{$ACC_{5}$ and $\overline{ACC}_{5}$ of PuriDivER with and without ODEDM across three runs with various buffer sizes on CIFAR10 under the standard and imbalanced settings. The best results in each setting are highlighted in \textbf{bold}.}
\begin{tabular}{@{}cccccc@{}}
\toprule
\multicolumn{2}{l}{\multirow{2}{*}{\textbf{Buffer Method}}} & \multicolumn{2}{c}{\textbf{Standard}}       & \multicolumn{2}{c}{\textbf{Imbalanced}}     \\
\multicolumn{2}{l}{}                                        & $ACC_{5}$        & $\overline{ACC}_{5}$ & $ACC_{5}$        & $\overline{ACC}_{5}$ \\ \midrule
\multirow{4}{*}{200}         & PuriDivER                    & $49.81\pm3.71$          & $61.55\pm15.25$  & $\mathbf{46.44}\pm1.59$ & $60.13\pm15.46$  \\
                             & PuriDivER (ODEDM-6)          & $\mathbf{50.12}\pm0.30$ & $62.95\pm13.62$  & $45.41\pm1.36$          & $58.58\pm14.92$  \\
                             & PuriDivER (ODEDM-13)         & $44.42\pm3.40$          & $59.06\pm15.32$  & $39.26\pm4.92$          & $55.15\pm16.16$  \\
                             & PuriDivER (ODEDM-19)         & $36.72\pm1.87$          & $56.73\pm18.43$  & $32.54\pm3.80$          & $51.68\pm19.18$  \\ \midrule
\multirow{4}{*}{500}         & PuriDivER                    & $57.29\pm4.31$          & $68.91\pm11.59$  & $\mathbf{61.55}\pm0.85$ & $69.35\pm13.22$  \\
                             & PuriDivER (ODEDM-16)         & $\mathbf{61.45}\pm0.84$ & $70.53\pm11.28$  & $58.15\pm0.68$          & $67.84\pm11.80$  \\
                             & PuriDivER (ODEDM-31)         & $55.70\pm1.48$          & $67.23\pm12.63$  & $55.31\pm1.66$          & $65.76\pm12.92$  \\
                             & PuriDivER (ODEDM-47)         & $46.28\pm4.57$          & $62.62\pm15.45$  & $39.75\pm10.21$         & $57.72\pm19.08$  \\ \bottomrule
\end{tabular}
\label{tab:PuriDivER}
\end{table*}

\subsection{Large-scale Dataset}
To assess performance on a large-scale benchmark, we report results under both the standard and imbalanced settings on ImageNet-R~\cite{hendrycks2021many}. The dataset is partitioned into 10 tasks, each containing 200 classes. We approximate the long-term memory proportion using the same ratio employed for TINYIMG, as shown in \cref{tab:k_approx}.

Given the increased input resolution of $224 \times 224$, we adopt the Vision Transformer (ViT)~\cite{dosovitskiy2021imageworth16x16words} instead of ResNet-18, as the latter typically exhibits suboptimal performance at this scale.

As shown in \cref{tab:imgnet-r}, the average accuracy remains extremely low (approximately 1.00\%) due to the intrinsic difficulty of the ImageNet-R benchmark. Nonetheless, meaningful distinctions emerge across methods and buffer sizes. With a small buffer of 200, ODEDM yields only marginal gains over DER\texttt{++}refresh, indicating that its effectiveness is constrained when memory is severely limited. In contrast, when the buffer size increases to 500, ODEDM exhibits more pronounced improvements, particularly in the standard $ACC_{10}$ setting, where ODEDM-1 attains 1.06\% compared to the baseline’s 0.75\%. A similar pattern is observed under the imbalanced $ACC_{10}$ configuration, with ODEDM improving performance to 1.00\% from 0.75\%, again surpassing DER\texttt{++}refresh baseline.

When the buffer is further expanded to 5120, ODEDM consistently delivers stronger performance, achieving the highest overall results: 1.14\% in $ACC_{10}$ under the standard setting and 1.06\% in $ACC_{10}$ under the imbalanced setting. Although the absolute gains are modest, an expected outcome given the difficulty of ImageNet-R, the improvements are stable and become increasingly pronounced as buffer capacity grows. In contrast, $\overline{ACC}_{10}$ performance either remains unchanged or deteriorates under both the standard and imbalanced conditions.

\begin{table*}[ht]
\scriptsize
\centering
\caption{$ACC_{10}$ and $\overline{ACC}_{10}$ of DER\texttt{++}refresh with and without ODEDM across three runs with various buffer sizes on ImageNet-R under the
standard and imbalanced settings. The best results in each setting are highlighted in \textbf{bold}.}
\begin{tabular}{@{}cccccc@{}}
\toprule
\multicolumn{2}{l}{\multirow{2}{*}{\textbf{Buffer Method}}} & \multicolumn{2}{c}{\textbf{Standard}}       & \multicolumn{2}{c}{\textbf{Imbalanced}} \\
\multicolumn{2}{l}{}                                        & $ACC_{10}$        & $\overline{ACC}_{10}$ & $ACC_{10}$        & $\overline{ACC}_{10}$  \\ \midrule
\multirow{2}{*}{200}    & DER\texttt{++}refresh             & $0.94\pm0.37$          & $2.08\pm1.71$  & $\mathbf{0.87}\pm0.28$     & $2.32\pm2.07$   \\
                        & DER\texttt{++}refresh (ODEDM-1)   & $\mathbf{0.95}\pm0.16$ & $4.94\pm7.33$  & $0.86\pm0.30$     & $2.15\pm1.60$   \\ \midrule
\multirow{3}{*}{500}    & DER\texttt{++}refresh             & $0.75\pm0.24$          & $2.57\pm2.18$  & $0.75\pm0.05$     & $2.15\pm2.06$   \\
                        & DER\texttt{++}refresh (ODEDM-1)   & $\mathbf{1.06}\pm0.25$ & $2.21\pm1.37$  & $\mathbf{1.00}\pm0.20$     & $2.18\pm1.80$   \\
                        & DER\texttt{++}refresh (ODEDM-2)   & $0.82\pm0.30$          & $2.41\pm2.47$  & $0.93\pm0.20$     & $2.66\pm2.63$   \\ \midrule
\multirow{4}{*}{5120}   & DER\texttt{++}refresh             & $0.78\pm0.13$          & $2.46\pm2.36$  & $0.64\pm0.04$     & $2.31\pm2.17$   \\
                        & DER\texttt{++}refresh (ODEDM-7)   & $0.99\pm0.12$          & $4.09\pm7.15$  & $0.84\pm0.07$     & $2.36\pm1.95$   \\
                        & DER\texttt{++}refresh (ODEDM-14)  & $\mathbf{1.14}\pm0.11$ & $2.50\pm1.88$  & $0.75\pm0.31$     & $2.51\pm2.03$   \\
                        & DER\texttt{++}refresh (ODEDM-21)  & $0.82\pm0.20$          & $2.17\pm1.44$  & $\mathbf{1.06}\pm0.16$     & $2.11\pm1.97$   \\ \bottomrule
\end{tabular}%
\label{tab:imgnet-r}
\end{table*}

\subsection{Other OCL Baselines}
We further compare DER\texttt{++}, with and without ODEDM, against eight established rehearsal-based OCL baselines on CIFAR10: Experience Replay (ER)~\cite{chaudhry2019tiny}, Gradient-based Sample Selection (GSS)~\cite{aljundi2019gradient}, Gradient Episodic Memory (GEM)~\cite{lopezpaz2022gradientepisodicmemorycontinual}, Averaged-GEM (A-GEM)~\cite{chaudhry2018efficient}, Maximally Interfered Retrieval (MIR)~\cite{aljundi2019online}, ASER~\cite{shim2021online}, CoPE~\cite{de2021continual}, and GDumb~\cite{prabhu2020gdumb}. For a stronger comparison, selected baselines, GEM, A-GEM, and FDR, are additionally trained in a full offline regime of 50 epochs.

ER mitigates catastrophic forgetting by uniformly sampling stored examples from the replay buffer. GSS extends ER by preferentially storing samples that maximise representational utility. GEM, and its computationally lighter variant A-GEM, regularise model updates by enforcing constraints based on gradients computed from buffer data. MIR prioritises replay samples according to their estimated interference with future losses. ASER employs Shapley value theory to guide buffer updates and to quantify the marginal contribution of individual samples. CoPE maintains momentum-updated class prototypes to encourage stable representations across tasks. GDumb greedily stores a class-balanced subset of incoming data and, at evaluation time, discards the original model and retrains a new one solely on the buffered samples.

As reported in Table~\ref{tab:onpro}, under the online training regime (1 epoch per task), A-GEM and GSS achieve the lowest $ACC_{5}$ values, remaining below 25\% across all buffer sizes. ER, MIR, and GDumb provide moderate improvements, with GDumb reaching 34.0\% when $M=500$. CoPE performs substantially better, achieving 42.9\% at $M=500$, while DER\texttt{++} reaches 39.9\% under the same configuration. When the baselines are extended to the offline setting (50 epochs), GEM and A-GEM exhibit only minor improvements, whereas FDR attains the strongest performance among them for buffer sizes of 200 and 500. In contrast, our method—DER\texttt{++} augmented with ODEDM—consistently achieves the highest $ACC_{5}$, obtaining 42.7\% with $M=200$ and 47.3\% with $M=500$.


\subsection{Ablation Study on Distance Metrics}
We conduct an ablation study to examine the effect of different distance metrics, namely, L1, L2, Maximum Mean Discrepancy (MMD), and Sinkhorn, on the performance of DER\texttt{++}refresh when integrated with ODEDM on CIFAR10. For each class, we sample 1,000 instances from the CIFAR10 training set and embed them into the model’s feature space, along with all buffer samples retained by ODEDM under a long-term memory proportion of 75\%. Principal Component Analysis (PCA) is applied to reduce these feature representations to two dimensions for visualisation.

To quantitatively assess the alignment between the feature space distributions of CIFAR10 and buffer samples, we fit a linear mapping to the projected features of each set and compute the cosine similarity between the resulting lines. In this analysis, the background fit refers to the linear mapping derived from CIFAR10 features, whereas the sample fit corresponds to that obtained from buffer features.

\begin{figure*}[ht]
    \centering
    \includegraphics[width=\linewidth]{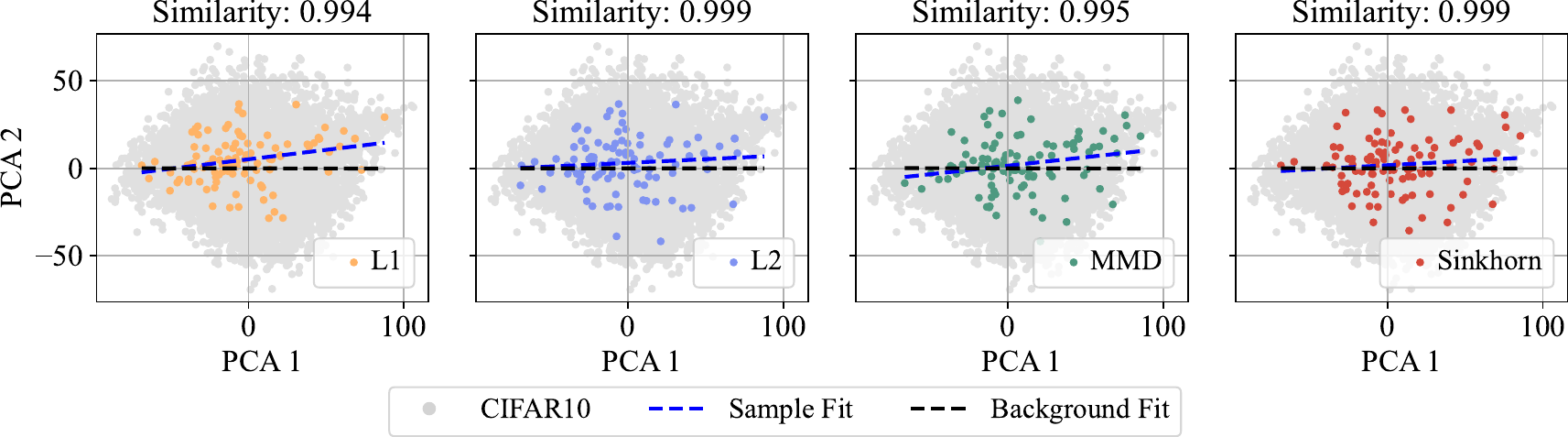}
    \caption{Visualisation of feature space in buffers and sampled CIFAR10 with PCA under different distance metrics in ODEDM.}
    \label{fig:pca}
\end{figure*}

As shown in \cref{fig:pca}, L2 and Sinkhorn distances achieve the highest cosine similarity values, indicating a closer alignment between the two distributions in the reduced feature space. These results suggest that L2 and Sinkhorn metrics more effectively capture the underlying structure of the data compared to L1 and MMD.

\subsection{Sample Visualisation}
\begin{figure*}[htbp]
  \centering
  \begin{minipage}{\linewidth}
    \centering
    \includegraphics[width=\linewidth]{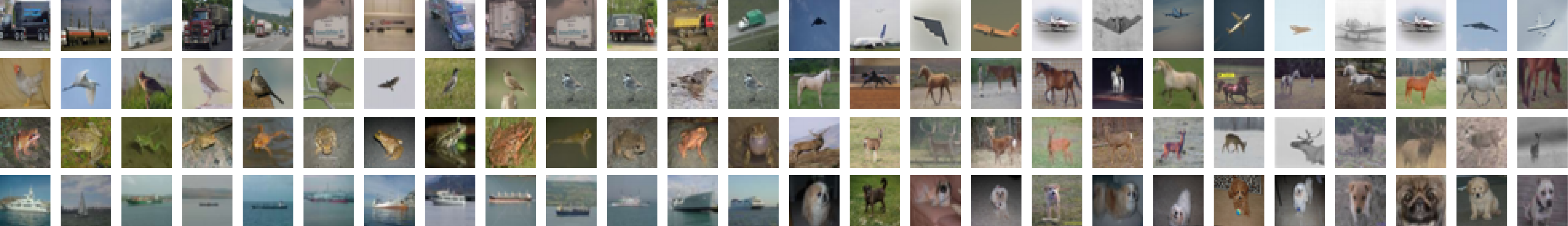}
    \subcaption{L1}\label{fig:l1_samples}
  \end{minipage}\vspace{1em}

  \begin{minipage}{\linewidth}
    \centering
    \includegraphics[width=\linewidth]{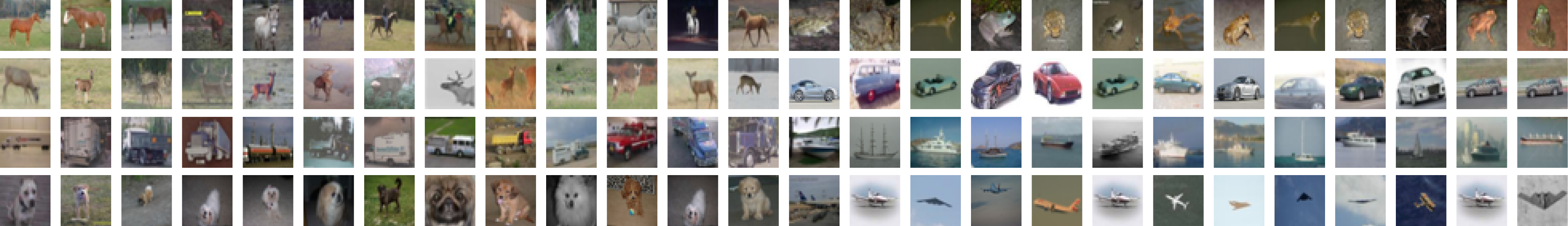}
    \subcaption{L2}\label{fig:l2_samples}
  \end{minipage}\vspace{1em}

  \begin{minipage}{\linewidth}
    \centering
    \includegraphics[width=\linewidth]{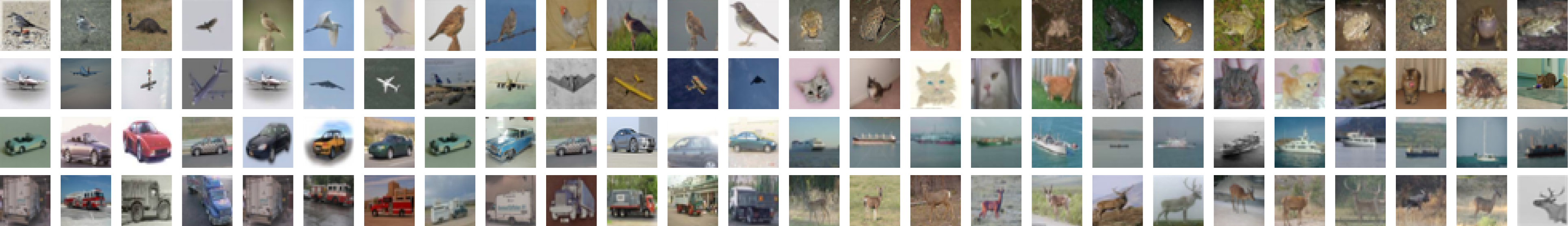}
    \subcaption{MMD}\label{fig:mmd_samples}
  \end{minipage}\vspace{1em}

  \begin{minipage}{\linewidth}
    \centering
    \includegraphics[width=\linewidth]{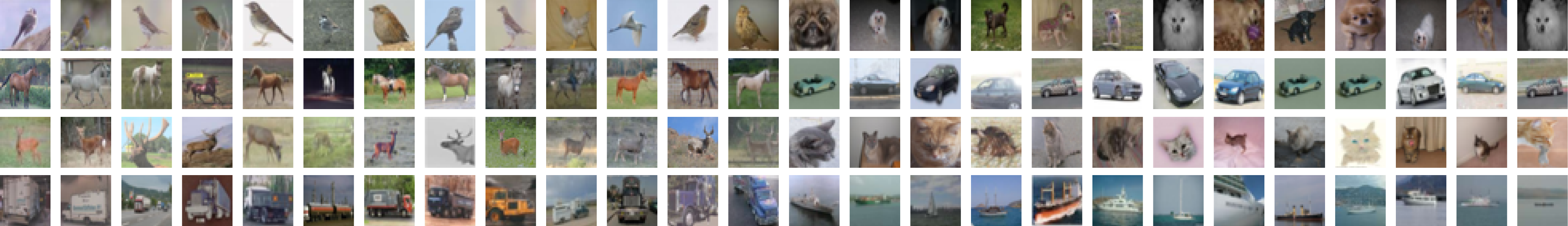}
    \subcaption{Sinkhorn}\label{fig:sinkhorn_samples}
  \end{minipage}

  \caption{Samples selected under different distance metrics on CIFAR10 under the 75\% long-term buffer proportion, with DER\texttt{++}refresh and ODEDM.}
  \label{fig:samples_all}
\end{figure*}

As shown in \cref{fig:samples_all}, we provide a visualisation of the samples selected by ODEDM under different distance metrics on CIFAR10. We use DER\texttt{++}refresh combined with ODEDM and a long-term memory buffer proportion of 75\%. For each distance metric, the rows correspond to successive tasks: the first row depicts the samples selected from Task 1, the second row from Task 2, and so on.

\subsection{Analysis on \texorpdfstring{$ACC_{T}^{\text{Task-IL}}$}{ACC\_T (Task-IL)}}

To complement the $ACC_T$ results in \cref{tab:accResults}, we further report $ACC_T^{\text{Task-IL}}$ on CIFAR10, CIFAR100 and TINYIMG in \cref{tab:cifar10_taskil_only,tab:cifar100_taskil_only,tab:tinyimg_taskil_only}. Across all datasets and buffer sizes, the $ACC_T^{\text{Task-IL}}$ accuracies are substantially higher than their $ACC_T$ counterparts, confirming that ODEDM also mitigates inter-task interference. On CIFAR10, most rehearsal-based methods with ODEDM achieve $ACC_5^{\text{Task-IL}}$ above 80\% even with a buffer of 200, and up to over 90\% when the buffer size is increased to 5120. For example, with a buffer of 5120, DER\texttt{++}refresh with ODEDM reaches $90.32\pm3.62$ under the standard setting, while VR-MCL and POCL attain $90.83\pm0.70$ and $92.60\pm1.96$, respectively. For the imbalanced setting, ODEDM continues to yield strong $ACC_5^{\text{Task-IL}}$ performance. For example, POCL combined with ODEDM attains an accuracy of $92.50\pm0.78$, showing that ODEDM effectively preserves inter-task separability even in the presence of substantial class-frequency skew.

On CIFAR100, $ACC_{10}^{\text{Task-IL}}$ follows a similar trend, increasing with buffer size and remaining consistently higher than the corresponding $ACC_{10}$ accuracies. With a buffer of 5120, DER\texttt{++}refresh with ODEDM achieves $63.24\pm0.85$ under the standard setting and $54.34\pm1.98$ in the imbalanced case, while VR-MCL and POCL further improve $ACC_{10}^{\text{Task-IL}}$ accuracy to $64.49\pm1.24$ and $72.64\pm1.17$, respectively. On TINYIMG, which is more challenging, the absolute values are lower, but the same pattern holds: with a buffer of 5120, DER\texttt{++}refresh with ODEDM reaches around $53$ $ACC_{10}^{\text{Task-IL}}$ accuracy, and both VR-MCL and POCL exceed $51$ across both standard and imbalanced settings. Overall, these results show that once task identity is provided, models equipped with ODEDM can retain strong per-task performance, and the performance gap between $ACC_{T}$ and $ACC_{T}^{\text{Task-IL}}$ metrics highlights that the main difficulty in OCL lies in handling label-space ambiguity across tasks rather than learning each task in isolation.

\clearpage

\begin{table*}[ht]
\scriptsize
\centering
\caption{$ACC_5^{\text{Task-IL}}$ of ODEDM on CIFAR10 across three runs under the standard and imbalanced settings with varying buffer sizes.}
\begin{tabular}{@{}cccc@{}}
\toprule
\multicolumn{2}{l}{\multirow{2}{*}{\textbf{Buffer Method}}} & \multicolumn{1}{c}{\textbf{Standard}} & \multicolumn{1}{c}{\textbf{Imbalanced}} \\
\multicolumn{2}{l}{}                                        & $ACC_5^{\text{Task-IL}}$              & $ACC_5^{\text{Task-IL}}$                \\ \midrule
\multirow{35}{*}{$200$} 
& DER                     & $\mathbf{81.25}\pm5.19$          & $\mathbf{73.61}\pm8.02$  \\
& DER (ODEDM-6)                     & $69.61\pm1.76$          & $57.80\pm7.82$  \\
& DER (ODEDM-13)                    & $75.25\pm2.30$ & $67.07\pm6.62$  \\
& DER (ODEDM-19)                    & $73.10\pm3.21$          & $59.28\pm7.96$  \\ \cmidrule(l){3-4}
& DER\texttt{++}                     & $\mathbf{82.60}\pm3.49$          & $\mathbf{82.80}\pm4.16$   \\
& DER\texttt{++} (ODEDM-6)          & $81.34\pm1.61$ & $82.27\pm1.65$  \\
& DER\texttt{++} (ODEDM-13)         & $79.53\pm4.29$          & $78.62\pm2.77$  \\
& DER\texttt{++} (ODEDM-19)         & $78.35\pm2.17$          & $76.22\pm3.25$  \\ \cmidrule(l){3-4}
& DER\texttt{++}refresh                     & $\mathbf{86.49}\pm3.04$          & $74.38\pm7.60$ 
  \\
& DER\texttt{++}refresh (ODEDM-6)   & $80.49\pm3.95$          & $77.69\pm2.72$  \\
& DER\texttt{++}refresh (ODEDM-13)  & $85.56\pm0.74$ & $\mathbf{79.79}\pm0.75$  \\
& DER\texttt{++}refresh (ODEDM-19)  & $82.68\pm1.14$          & $79.20\pm1.11$  \\ \cmidrule(l){3-4}
& FDR                     & $69.93\pm6.02$          & $\mathbf{69.37}\pm8.89$  \\
& FDR (ODEDM-6)                     & $67.12\pm6.33$          & $68.20\pm3.27$  \\
& FDR (ODEDM-13)                    & $66.31\pm4.40$          & $63.77\pm2.12$  \\
& FDR (ODEDM-19)                    & $\mathbf{73.76}\pm5.98$ & $63.70\pm6.24$  \\ \cmidrule(l){3-4}
& iCaRL                     & $88.99\pm0.85$          & $\mathbf{84.85}\pm4.48$   \\
& iCaRL (ODEDM-6)                   & $\mathbf{89.88}\pm1.72$ & $81.94\pm5.76$  \\
& iCaRL (ODEDM-13)                  & $85.40\pm3.42$          & $80.78\pm5.69$  \\
& iCaRL (ODEDM-19)                  & $83.45\pm2.96$          & $81.57\pm4.09$  \\ \cmidrule(l){3-4}
& PuriDivER                     & $\mathbf{79.93}\pm0.92$          & $\mathbf{77.75}\pm2.31$  \\
& PuriDivER (ODEDM-6)               & $78.70\pm0.58$ & $75.23\pm1.07$  \\
& PuriDivER (ODEDM-13)              & $78.68\pm0.83$          & $76.28\pm0.44$  \\
& PuriDivER (ODEDM-19)              & $77.47\pm0.48$          & $76.43\pm2.32$  \\ \cmidrule(l){3-4}
& VR-MCL                     & $85.58\pm2.58$          & $80.00\pm4.81$  \\
& VR-MCL (ODEDM-6)                  & $82.77\pm2.30$          & $\mathbf{86.19}\pm1.51$  \\
& VR-MCL (ODEDM-13)                 & $81.97\pm6.35$          & $83.12\pm3.14$  \\
& VR-MCL (ODEDM-19)                 & $\mathbf{86.52}\pm2.68$ & $82.61\pm1.20$  \\ \cmidrule(l){3-4}
& POCL                     & $80.67\pm4.70$          & $78.95\pm3.08$  \\
& POCL (ODEDM-6)                    & $\mathbf{81.99}\pm3.46$ & $78.32\pm2.61$  \\
& POCL (ODEDM-13)                   & $76.09\pm3.07$          & $71.80\pm9.26$  \\
& POCL (ODEDM-19)                   & $80.88\pm2.23$          & $\mathbf{80.51}\pm0.86$  \\ \midrule
\multirow{35}{*}{$500$}
& DER                     & $\mathbf{85.74}\pm3.29$           & $\mathbf{82.62}\pm2.78$   \\
& DER (ODEDM-16)                    & $75.26\pm5.04$          & $61.48\pm4.06$  \\
& DER (ODEDM-31)                    & $70.96\pm3.70$          & $70.92\pm3.97$  \\
& DER (ODEDM-47)                    & $72.68\pm5.20$          & $62.80\pm6.85$  \\ \cmidrule(l){3-4}
& DER\texttt{++}                     & $\mathbf{86.08}\pm4.85$          & $82.21\pm4.42$   \\
& DER\texttt{++} (ODEDM-16)         & $83.46\pm1.02$          & $\mathbf{82.85}\pm0.65$  \\
& DER\texttt{++} (ODEDM-31)         & $83.05\pm1.19$          & $76.59\pm2.68$  \\
& DER\texttt{++} (ODEDM-47)         & $83.37\pm0.19$          & $75.66\pm1.49$  \\ \cmidrule(l){3-4}
& DER\texttt{++}refresh                 & $\mathbf{85.71}\pm5.02$          & $\mathbf{82.31}\pm5.58$   \\
& DER\texttt{++}refresh (ODEDM-16)  & $84.77\pm2.58$          & $72.08\pm2.30$  \\
& DER\texttt{++}refresh (ODEDM-31)  & $81.19\pm4.23$          & $77.49\pm5.68$  \\
& DER\texttt{++}refresh (ODEDM-47)  & $83.14\pm1.60$          & $79.47\pm3.30$  \\ \cmidrule(l){3-4}
& FDR                 & $68.27\pm4.50$          & $\mathbf{73.79}\pm4.25$  \\
& FDR (ODEDM-16)                    & $\mathbf{77.52}\pm2.14$          & $67.96\pm2.47$  \\
& FDR (ODEDM-31)                    & $68.20\pm1.74$          & $65.12\pm4.38$  \\
& FDR (ODEDM-47)                    & $64.49\pm6.19$          & $59.64\pm2.36$  \\ \cmidrule(l){3-4}
& iCaRL                 & $84.89\pm5.17$           & $\mathbf{87.46}\pm1.57$   \\
& iCaRL (ODEDM-16)                  & $80.84\pm7.76$          & $85.26\pm2.64$  \\
& iCaRL (ODEDM-31)                  & $\mathbf{86.81}\pm3.38$ & $80.89\pm1.93$  \\
& iCaRL (ODEDM-47)                  & $86.61\pm2.97$          & $85.30\pm2.96$  \\ \cmidrule(l){3-4}
& PuriDivER                 & $80.40\pm0.38$          & $\mathbf{79.86}\pm0.08$  \\
& PuriDivER (ODEDM-16)              & $82.49\pm0.92$          & $78.93\pm0.54$  \\
& PuriDivER (ODEDM-31)              & $\mathbf{83.37}\pm1.49$          & $78.00\pm2.06$  \\
& PuriDivER (ODEDM-47)              & $80.01\pm0.60$          & $78.29\pm0.86$  \\ \cmidrule(l){3-4}
& VR-MCL                 & $87.89\pm2.14$          & $82.28\pm1.09$  \\
& VR-MCL (ODEDM-16)                 & $\mathbf{88.66}\pm0.29$          & $\mathbf{88.18}\pm0.20$  \\
& VR-MCL (ODEDM-31)                 & $87.24\pm3.69$          & $84.59\pm1.94$  \\
& VR-MCL (ODEDM-47)                 & $84.74\pm1.35$          & $83.51\pm2.36$  \\ \cmidrule(l){3-4}
& POCL                 & $83.42\pm1.49$          & $79.07\pm6.93$  \\
& POCL (ODEDM-16)                   & $\mathbf{87.09}\pm0.90$          & $\mathbf{82.89}\pm2.24$  \\
& POCL (ODEDM-31)                   & $85.61\pm3.78$          & $82.56\pm2.21$  \\
& POCL (ODEDM-47)                   & $80.94\pm5.17$          & $77.69\pm6.66$  \\ \bottomrule
\end{tabular}%
\label{tab:cifar10_taskil_only}
\end{table*}

\begin{table*}[ht]
\scriptsize
\centering
\begin{tabular}{@{}cccc@{}}
\toprule
\multicolumn{2}{l}{\multirow{2}{*}{\textbf{Buffer Method}}} & \multicolumn{1}{c}{\textbf{Standard}} & \multicolumn{1}{c}{\textbf{Imbalanced}} \\
\multicolumn{2}{l}{}                                        & $ACC_5^{\text{Task-IL}}$              & $ACC_5^{\text{Task-IL}}$                \\ \midrule
\multirow{30}{*}{$5120$}
& DER                 & $\mathbf{84.07}\pm3.55$          & $\mathbf{74.88}\pm6.59$   \\
& DER (ODEDM-160)                   & $66.60\pm3.06$          & $58.35\pm3.24$  \\
& DER (ODEDM-320)                   & $66.24\pm2.21$          & $63.16\pm2.46$  \\
& DER (ODEDM-480)                   & $76.19\pm3.84$          & $65.31\pm1.93$  \\ \cmidrule(l){3-4}
& DER\texttt{++}                 & $86.26\pm1.41$          & $83.81\pm4.33$   \\
& DER\texttt{++} (ODEDM-160)        & $\mathbf{87.77}\pm3.28$          & $81.29\pm4.82$  \\
& DER\texttt{++} (ODEDM-320)        & $85.95\pm3.53$          & $86.64\pm1.70$  \\
& DER\texttt{++} (ODEDM-480)        & $84.18\pm8.06$          & $\mathbf{87.37}\pm2.42$  \\ \cmidrule(l){3-4}
& DER\texttt{++}refresh                 & $88.64\pm2.79$          & $81.72\pm2.88$   \\
& DER\texttt{++}refresh (ODEDM-160) & $87.82\pm1.57$          & $\mathbf{84.50}\pm4.13$  \\
& DER\texttt{++}refresh (ODEDM-320) & $\mathbf{90.32}\pm3.62$ & $77.80\pm1.75$  \\
& DER\texttt{++}refresh (ODEDM-480) & $87.70\pm1.89$          & $81.12\pm3.78$  \\ \cmidrule(l){3-4}
& FDR                 & $\mathbf{79.53}\pm6.25$          & $66.57\pm3.28$  \\
& FDR (ODEDM-160)                   & $64.69\pm6.23$          & $\mathbf{69.99}\pm3.47$  \\
& FDR (ODEDM-320)                   & $64.53\pm4.79$          & $61.47\pm3.36$  \\
& FDR (ODEDM-480)                   & $67.21\pm5.14$          & $63.00\pm0.00$  \\ \cmidrule(l){3-4}
& iCaRL                 & $86.20\pm3.43$          & $82.29\pm6.68$  \\
& iCaRL (ODEDM-160)                 & $\mathbf{90.27}\pm1.56$          & $85.40\pm1.81$  \\
& iCaRL (ODEDM-320)                 & $88.27\pm1.63$          & $\mathbf{87.87}\pm1.47$  \\
& iCaRL (ODEDM-480)                 & $84.29\pm2.60$          & $84.39\pm2.71$  \\ \cmidrule(l){3-4}
& VR-MCL                 & $88.32\pm2.58$          & $87.21\pm2.44$  \\
& VR-MCL (ODEDM-160)                & $90.59\pm0.72$          & $\mathbf{90.68}\pm3.08$  \\
& VR-MCL (ODEDM-320)                & $\mathbf{90.83}\pm0.70$          & $87.44\pm0.79$  \\
& VR-MCL (ODEDM-480)                & $90.74\pm2.61$          & $89.09\pm2.06$  \\ \cmidrule(l){3-4}
& POCL                 & $91.42\pm1.63$          & $91.03\pm0.76$  \\
& POCL (ODEDM-160)                  & $92.11\pm1.73$          & $91.25\pm2.45$  \\
& POCL (ODEDM-320)                  & $\mathbf{92.60}\pm1.96$ & $88.05\pm1.78$  \\
& POCL (ODEDM-480)                  & $91.80\pm1.40$          & $\mathbf{92.50}\pm0.78$  \\ \bottomrule
\end{tabular}%
\end{table*}

\begin{table*}[ht]
\scriptsize
\centering
\caption{$ACC_{10}^{\text{Task-IL}}$ of ODEDM on CIFAR100 across three runs under the standard and imbalanced settings with varying buffer sizes.}
\begin{tabular}{@{}cccc@{}}
\toprule
\multicolumn{2}{l}{\multirow{2}{*}{\textbf{Buffer Method}}} & \multicolumn{1}{c}{\textbf{Standard}} & \multicolumn{1}{c}{\textbf{Imbalanced}} \\
\multicolumn{2}{l}{}                                        & $ACC_{10}^{\text{Task-IL}}$           & $ACC_{10}^{\text{Task-IL}}$             \\ \midrule
\multirow{20}{*}{$200$} 
& DER                 & $\mathbf{34.00}\pm1.67$          & $\mathbf{30.37}\pm1.87$  \\
& DER (ODEDM-1)                    & $25.53\pm2.25$          & $20.77\pm2.63$  \\
\cmidrule(l){3-4}
& DER\texttt{++}                 & $\mathbf{41.51}\pm2.68$          & $36.52\pm1.08$  \\
& DER\texttt{++} (ODEDM-1)         & $38.27\pm4.86$          & $\mathbf{38.66}\pm3.79$  \\
\cmidrule(l){3-4}
& DER\texttt{++}refresh                 & $\mathbf{41.26}\pm0.79$          & $\mathbf{40.30}\pm3.94$  \\
& DER\texttt{++}refresh (ODEDM-1)  & $41.17\pm0.87$ & $34.87\pm5.62$  \\
\cmidrule(l){3-4}
& FDR                 & $\mathbf{34.71}\pm2.53$          & $\mathbf{27.84}\pm2.44$  \\
& FDR (ODEDM-1)                    & $23.10\pm3.18$          & $21.11\pm3.55$  \\
\cmidrule(l){3-4}
& iCaRL                 & $\mathbf{35.59}\pm0.58$          & $\mathbf{33.79}\pm0.61$   \\
& iCaRL (ODEDM-1)                  & $7.05\pm1.19$           & $6.28\pm1.33$  \\
\cmidrule(l){3-4}
& PuriDivER                 & $\mathbf{37.39}\pm1.25$          & $\mathbf{36.63}\pm0.56$  \\
& PuriDivER (ODEDM-1)              & $36.63\pm0.56$          & $35.18\pm0.35$  \\
\cmidrule(l){3-4}
& VR-MCL                 & $45.68\pm0.38$          & $40.71\pm0.36$  \\
& VR-MCL (ODEDM-1)                 & $\mathbf{51.53}\pm1.11$          & $\mathbf{47.14}\pm0.95$  \\
\cmidrule(l){3-4}
& POCL                 & $39.94\pm2.02$          & $38.21\pm1.14$  \\
& POCL (ODEDM-1)                   & $\mathbf{43.41}\pm0.95$          & $\mathbf{40.88}\pm1.55$  \\ \midrule
\multirow{35}{*}{$500$}
& DER                 & $\mathbf{33.45}\pm3.36$         & $\mathbf{30.81}\pm0.40$  \\
& DER (ODEDM-1)                    & $24.88\pm1.12$          & $21.33\pm3.94$  \\
& DER (ODEDM-3)                    & $22.15\pm1.55$          & $22.21\pm1.70$  \\
& DER (ODEDM-4)                    & $22.31\pm3.58$          & $23.59\pm1.54$  \\ \cmidrule(l){3-4}
& DER\texttt{++}                 & $49.41\pm2.91$           & $41.87\pm6.81$   \\
& DER\texttt{++} (ODEDM-1)         & $45.26\pm6.63$          & $\mathbf{44.82}\pm2.34$  \\
& DER\texttt{++} (ODEDM-3)         & $\mathbf{50.26}\pm2.38$ & $43.40\pm1.46$  \\
& DER\texttt{++} (ODEDM-4)         & $47.70\pm0.56$          & $43.26\pm2.48$  \\ \cmidrule(l){3-4}
& DER\texttt{++}refresh                 & $\mathbf{54.13}\pm1.85$          & $45.27\pm1.73$  \\
& DER\texttt{++}refresh (ODEDM-1)  & $44.83\pm4.61$          & $\mathbf{45.69}\pm2.42$  \\
& DER\texttt{++}refresh (ODEDM-3)  & $46.30\pm1.57$          & $43.89\pm3.38$  \\
& DER\texttt{++}refresh (ODEDM-4)  & $47.21\pm0.63$          & $42.65\pm4.40$  \\ \cmidrule(l){3-4}
& FDR                 & $\mathbf{32.71}\pm3.47$          & $\mathbf{27.94}\pm4.77$   \\
& FDR (ODEDM-1)                    & $25.51\pm2.24$          & $25.18\pm1.66$  \\
& FDR (ODEDM-3)                    & $29.50\pm2.47$          & $22.23\pm4.34$  \\
& FDR (ODEDM-4)                    & $28.67\pm1.96$          & $25.86\pm1.47$  \\ \cmidrule(l){3-4}
& iCaRL                 & $\mathbf{36.81}\pm0.94$          & $\mathbf{34.53}\pm0.47$   \\
& iCaRL (ODEDM-1)                  & $33.70\pm0.45$          & $32.81\pm1.27$  \\
& iCaRL (ODEDM-3)                  & $8.75\pm0.66$           & $9.29\pm0.69$  \\
& iCaRL (ODEDM-4)                  & $8.28\pm0.96$           & $8.91\pm0.68$  \\ \cmidrule(l){3-4}
& PuriDivER                 & $42.22\pm1.18$           & $38.93\pm1.29$  \\
& PuriDivER (ODEDM-1)              & $41.99\pm0.46$          & $39.89\pm1.38$  \\
& PuriDivER (ODEDM-3)              & $\mathbf{44.72}\pm1.35$          & $\mathbf{40.30}\pm0.58$  \\
& PuriDivER (ODEDM-4)              & $40.55\pm1.53$          & $37.24\pm1.96$  \\ \cmidrule(l){3-4}
& VR-MCL                 & $53.12\pm1.10$          & $48.05\pm1.54$  \\
& VR-MCL (ODEDM-1)                 & $54.60\pm1.51$          & $51.43\pm1.46$  \\
& VR-MCL (ODEDM-3)                 & $\mathbf{55.79}\pm0.89$          & $\mathbf{51.93}\pm0.85$  \\
& VR-MCL (ODEDM-4)                 & $55.68\pm0.63$          & $51.61\pm1.45$  \\ \cmidrule(l){3-4}
& POCL                 & $46.13\pm1.17$          & $44.07\pm0.67$  \\
& POCL (ODEDM-1)                   & $\mathbf{52.96}\pm2.51$          & $\mathbf{47.96}\pm1.30$  \\
& POCL (ODEDM-3)                   & $49.09\pm0.56$          & $47.06\pm2.41$  \\
& POCL (ODEDM-4)                   & $45.04\pm1.97$          & $44.76\pm0.93$  \\
\bottomrule
\end{tabular}
\label{tab:cifar100_taskil_only}
\end{table*}

\begin{table*}[ht]
\scriptsize
\centering
\begin{tabular}{@{}cccc@{}}
\toprule
\multicolumn{2}{l}{\multirow{2}{*}{\textbf{Buffer Method}}} & \multicolumn{1}{c}{\textbf{Standard}} & \multicolumn{1}{c}{\textbf{Imbalanced}} \\
\multicolumn{2}{l}{}                                        & $ACC_{10}^{\text{Task-IL}}$           & $ACC_{10}^{\text{Task-IL}}$             \\ \midrule
\multirow{30}{*}{$5120$}
& DER                     & $\mathbf{31.26}\pm2.15$          & $\mathbf{31.32}\pm1.52$   \\
& DER (ODEDM-14)                   & $24.25\pm2.44$          & $18.84\pm2.60$  \\
& DER (ODEDM-28)                   & $26.29\pm1.48$          & $20.76\pm2.96$  \\
& DER (ODEDM-42)                   & $23.44\pm2.05$          & $18.54\pm4.31$  \\ \cmidrule(l){3-4}
& DER\texttt{++}               & $57.62\pm5.17$           & $46.23\pm10.71$  \\
& DER\texttt{++} (ODEDM-14)        & $\mathbf{61.76}\pm5.09$          & $50.53\pm7.10$  \\
& DER\texttt{++} (ODEDM-28)        & $58.53\pm3.26$          & $\mathbf{53.41}\pm2.51$  \\
& DER\texttt{++} (ODEDM-42)        & $57.57\pm3.07$          & $53.39\pm1.38$  \\ \cmidrule(l){3-4}
& DER\texttt{++}refresh               & $53.64\pm2.07$          & $39.67\pm7.60$  \\
& DER\texttt{++}refresh (ODEDM-14) & $56.36\pm5.40$          & $52.73\pm2.98$  \\
& DER\texttt{++}refresh (ODEDM-28) & $\mathbf{63.24}\pm0.85$ & $\mathbf{54.34}\pm1.98$  \\
& DER\texttt{++}refresh (ODEDM-42) & $59.90\pm3.95$          & $47.47\pm4.97$  \\ \cmidrule(l){3-4}
& FDR               & $36.85\pm0.76$         & $\mathbf{33.16}\pm2.56$  \\
& FDR (ODEDM-14)                   & $29.69\pm6.18$          & $27.68\pm1.61$  \\
& FDR (ODEDM-28)                   & $\mathbf{30.30}\pm2.69$          & $25.52\pm1.41$  \\
& FDR (ODEDM-42)                   & $29.42\pm2.04$          & $28.15\pm0.63$  \\ \cmidrule(l){3-4}
& iCaRL               & $\mathbf{37.47}\pm0.97$          & $34.45\pm0.74$  \\
& iCaRL (ODEDM-14)                 & $35.42\pm0.67$          & $33.80\pm1.77$  \\
& iCaRL (ODEDM-28)                 & $36.51\pm0.54$          & $\mathbf{35.38}\pm0.44$  \\
& iCaRL (ODEDM-42)                 & $35.63\pm0.96$          & $33.85\pm0.67$  \\ \cmidrule(l){3-4}
& VR-MCL               & $59.44\pm0.43$          & $53.34\pm0.99$  \\
& VR-MCL (ODEDM-14)                & $\mathbf{64.49}\pm1.24$          & $59.74\pm0.61$  \\
& VR-MCL (ODEDM-28)                & $63.90\pm1.82$          & $58.90\pm1.14$  \\
& VR-MCL (ODEDM-42)                & $63.26\pm1.68$          & $\mathbf{60.72}\pm0.58$  \\ \cmidrule(l){3-4}
& POCL               & $65.03\pm1.20$          & $\mathbf{65.03}\pm1.20$  \\
& POCL (ODEDM-14)                  & $70.36\pm0.95$          & $64.86\pm3.34$  \\
& POCL (ODEDM-28)                  & $69.75\pm1.12$          & $61.42\pm4.18$  \\
& POCL (ODEDM-42)                  & $\mathbf{72.64}\pm1.17$ & $64.63\pm2.12$  \\ \bottomrule
\end{tabular}
\end{table*}

\begin{table*}[ht]
\scriptsize
\centering
\caption{$ACC_{10}^{\text{Task-IL}}$ of ODEDM on TINYIMG across three runs under the standard and imbalanced settings with varying buffer sizes.}
\begin{tabular}{@{}cccc@{}}
\toprule
\multicolumn{2}{l}{\multirow{2}{*}{\textbf{Buffer Method}}} & \multicolumn{1}{c}{\textbf{Standard}} & \multicolumn{1}{c}{\textbf{Imbalanced}} \\
\multicolumn{2}{l}{}                                        & $ACC_{10}^{\text{Task-IL}}$           & $ACC_{10}^{\text{Task-IL}}$             \\ \midrule
\multirow{17}{*}{$200$}
& DER               & $\mathbf{27.15}\pm3.47$         & $\mathbf{20.86}\pm1.75$  \\
& DER (ODEDM-1)                    & $17.21\pm0.29$          & $14.44\pm1.40$  \\
\cmidrule(l){3-4}
& DER\texttt{++}               & $\mathbf{31.25}\pm1.20$           & $\mathbf{28.56}\pm2.28$   \\
& DER\texttt{++} (ODEDM-1)         & $27.17\pm2.65$          & $26.70\pm2.11$  \\
\cmidrule(l){3-4}
& DER\texttt{++}refresh               & $\mathbf{34.15}\pm0.96$          & $\mathbf{27.10}\pm0.53$ \\
& DER\texttt{++}refresh (ODEDM-1)  & $28.04\pm0.88$ & $23.28\pm3.68$  \\
\cmidrule(l){3-4}
& FDR               & $\mathbf{25.32}\pm2.31$        & $\mathbf{19.50}\pm1.77$  \\
& FDR (ODEDM-1)                    & $18.65\pm3.00$          & $15.70\pm2.60$  \\
\cmidrule(l){3-4}
& iCaRL               & $\mathbf{19.16}\pm0.52$        & $\mathbf{17.82}\pm0.48$  \\
& iCaRL (ODEDM-1)                  & $-$                     & $-$  \\
\cmidrule(l){3-4}
& VR-MCL               & $31.67\pm0.56$          & $29.23\pm0.72$  \\
& VR-MCL (ODEDM-1)                 & $\mathbf{36.62}\pm0.77$          & $\mathbf{32.91}\pm0.93$  \\
\cmidrule(l){3-4}
& POCL               & $25.81\pm2.63$          & $\mathbf{24.87}\pm0.46$  \\
& POCL (ODEDM-1)                   & $\mathbf{28.01}\pm2.41$          & $23.51\pm1.78$  \\ \midrule
\multirow{23}{*}{$500$}
& DER                 & $\mathbf{26.34}\pm1.34$        & $\mathbf{22.07}\pm1.36$ \\
& DER (ODEDM-1)                    & $20.52\pm1.00$          & $18.28\pm0.57$  \\
& DER (ODEDM-2)                    & $18.97\pm1.33$          & $15.14\pm0.23$  \\ \cmidrule(l){3-4}
& DER\texttt{++}                 & $\mathbf{39.53}\pm3.22$         & $\mathbf{33.92}\pm2.55$  \\
& DER\texttt{++} (ODEDM-1)         & $34.96\pm4.37$          & $31.60\pm1.65$  \\
& DER\texttt{++} (ODEDM-2)         & $35.92\pm3.48$ & $32.56\pm1.73$  \\ \cmidrule(l){3-4}
& DER\texttt{++}refresh              & $\mathbf{39.53}\pm1.38$        & $32.95\pm1.99$ \\
& DER\texttt{++}refresh (ODEDM-1)  & $35.66\pm1.90$          & $\mathbf{33.13}\pm1.75$  \\
& DER\texttt{++}refresh (ODEDM-2)  & $33.95\pm2.00$          & $31.74\pm1.21$  \\ \cmidrule(l){3-4}
& FDR                 & $\mathbf{22.54}\pm4.97$       & $\mathbf{19.15}\pm1.85$  \\
& FDR (ODEDM-1)                    & $21.14\pm1.48$          & $17.23\pm1.01$  \\
& FDR (ODEDM-2)                    & $20.33\pm1.46$          & $17.04\pm1.90$  \\ \cmidrule(l){3-4}
& iCaRL                 & $\mathbf{20.09}\pm0.56$        & $\mathbf{18.59}\pm0.93$  \\
& iCaRL (ODEDM-1)                  & $3.91\pm0.16$           & $3.46\pm0.12$  \\
& iCaRL (ODEDM-2)                  & $3.59\pm0.14$           & $3.62\pm0.39$  \\ \cmidrule(l){3-4}
& VR-MCL                 & $37.64\pm1.12$          & $35.48\pm0.53$  \\
& VR-MCL (ODEDM-1)                 & $39.86\pm0.56$          & $\mathbf{38.77}\pm2.13$  \\
& VR-MCL (ODEDM-2)                 & $\mathbf{40.13}\pm1.42$          & $36.69\pm1.54$  \\ \cmidrule(l){3-4}
& POCL                 & $28.77\pm3.47$          & $30.35\pm2.05$  \\
& POCL (ODEDM-1)                   & $\mathbf{34.74}\pm0.59$          & $\mathbf{30.86}\pm0.66$  \\
& POCL (ODEDM-2)                   & $32.69\pm1.71$          & $28.69\pm0.87$  \\
\bottomrule
\end{tabular}%
\label{tab:tinyimg_taskil_only}
\end{table*}

\begin{table*}[ht]
\scriptsize
\centering
\begin{tabular}{@{}cccc@{}}
\toprule
\multicolumn{2}{l}{\multirow{2}{*}{\textbf{Buffer Method}}} & \multicolumn{1}{c}{\textbf{Standard}} & \multicolumn{1}{c}{\textbf{Imbalanced}} \\
\multicolumn{2}{l}{}                                        & $ACC_{10}^{\text{Task-IL}}$           & $ACC_{10}^{\text{Task-IL}}$             \\ \midrule
\multirow{30}{*}{$5120$} 
& DER                 & $\mathbf{30.55}\pm3.65$        & $\mathbf{26.81}\pm0.97$ \\
& DER (ODEDM-7)                    & $20.62\pm0.84$          & $17.35\pm0.86$  \\
& DER (ODEDM-14)                   & $21.07\pm0.78$          & $18.27\pm1.30$  \\
& DER (ODEDM-21)                   & $20.28\pm0.52$          & $16.13\pm0.36$  \\ \cmidrule(l){3-4}
& DER\texttt{++}                 & $51.00\pm3.69$      & $46.06\pm1.88$ \\
& DER\texttt{++} (ODEDM-7)         & $\mathbf{51.29}\pm3.23$          & $42.29\pm3.17$  \\
& DER\texttt{++} (ODEDM-14)        & $48.58\pm2.44$          & $45.89\pm3.28$  \\
& DER\texttt{++} (ODEDM-21)        & $51.02\pm1.45$          & $\mathbf{47.35}\pm1.18$  \\ \cmidrule(l){3-4}
& DER\texttt{++}refresh            & $50.38\pm2.32$       & $43.35\pm0.92$  \\
& DER\texttt{++}refresh (ODEDM-7)  & $53.08\pm0.80$          & $\mathbf{47.61}\pm2.07$  \\
& DER\texttt{++}refresh (ODEDM-14) & $52.50\pm0.91$          & $44.53\pm3.45$  \\
& DER\texttt{++}refresh (ODEDM-21) & $\mathbf{53.27}\pm2.02$ & $45.85\pm1.39$  \\ \cmidrule(l){3-4}
& FDR                 & $22.44\pm1.29$       & $\mathbf{21.73}\pm0.64$  \\
& FDR (ODEDM-7)                    & $22.22\pm2.69$          & $21.09\pm1.93$  \\
& FDR (ODEDM-14)                   & $21.82\pm0.36$          & $20.66\pm0.26$  \\
& FDR (ODEDM-21)                   & $\mathbf{23.18}\pm2.81$          & $20.94\pm1.30$  \\ \cmidrule(l){3-4}
& iCaRL                 & $\mathbf{20.45}\pm1.64$       & $\mathbf{19.17}\pm0.98$  \\
& iCaRL (ODEDM-7)                  & $19.25\pm0.49$          & $18.23\pm0.47$  \\
& iCaRL (ODEDM-14)                 & $19.10\pm0.19$          & $17.88\pm0.22$  \\
& iCaRL (ODEDM-21)                 & $18.39\pm0.64$          & $17.45\pm0.49$  \\ \cmidrule(l){3-4}
& VR-MCL                 & $49.38\pm0.91$          & $44.27\pm0.91$  \\
& VR-MCL (ODEDM-7)                 & $51.70\pm1.08$          & $47.97\pm0.98$  \\
& VR-MCL (ODEDM-14)                & $51.91\pm0.35$          & $\mathbf{49.98}\pm0.82$  \\
& VR-MCL (ODEDM-21)                & $\mathbf{53.51}\pm0.53$          & $49.08\pm1.11$  \\ \cmidrule(l){3-4}
& POCL                 & $49.66\pm0.96$          & $44.96\pm1.12$  \\
& POCL (ODEDM-7)                   & $52.40\pm0.64$          & $49.87\pm0.40$  \\
& POCL (ODEDM-14)                  & $52.53\pm1.08$          & $50.01\pm1.90$  \\
& POCL (ODEDM-21)                  & $\mathbf{52.76}\pm0.58$          & $\mathbf{51.52}\pm0.73$  \\ \bottomrule
\end{tabular}%
\end{table*}


\end{document}